\documentclass[10pt,letterpaper]{article}
\usepackage[top=0.85in,left=2.75in,footskip=0.75in]{geometry}

\usepackage{comment}
\usepackage{graphicx}
\graphicspath{{./Figure/}}
\usepackage{epstopdf}
\AppendGraphicsExtensions{.tif}
\usepackage{amsmath,amssymb}

\usepackage{changepage}

\usepackage{textcomp,marvosym}

\usepackage{cite}

\usepackage{nameref,hyperref}

\usepackage[right]{lineno}

\usepackage[nopatch=eqnum]{microtype}
\DisableLigatures[f]{encoding = *, family = * }

\usepackage[table]{xcolor}

\usepackage{array}

\newcolumntype{+}{!{\vrule width 2pt}}

\newlength\savedwidth

\raggedright
\usepackage[aboveskip=1pt,labelfont=bf,labelsep=period,justification=raggedright,singlelinecheck=off]{caption}

\makeatletter
\renewcommand{\@biblabel}[1]{\quad#1.}
\makeatother

\usepackage{lastpage,fancyhdr,graphicx}
\usepackage{epstopdf}
\fancyheadoffset[L]{2.25in}
\fancyfootoffset[L]{2.25in}
\usepackage{CJKutf8}
\usepackage{booktabs} % 表をきれいにする
\usepackage{siunitx}  % 単位表記を統一する
\usepackage{pifont} % チェックマーク用
\usepackage{booktabs}
\usepackage{tabularx}

\usepackage{subcaption}
\begin{document}

%%あとで消す
% \begin{CJK}{UTF8}{min}

\vspace*{0.2in}

\begin{flushleft}
{\Large
\textbf\newline{MOFU: Development of a MOrphing Fluffy Unit with expansion and contraction capabilities and evaluation of the animacy of its movements} %(134 characters) Please use "sentence case" for title and headings (capitalize only the first word in a title (or heading), the first word in a subtitle (or subheading), and any proper nouns).
}
\newline
\\
Taisei Mogi\textsuperscript{1},
Mari Saito\textsuperscript{2},
Yoshihiro Nakata\textsuperscript{1}*
% Name1 Surname\textsuperscript{1,2\Yinyang},
% Name2 Surname\textsuperscript{2\Yinyang},
% Name3 Surname\textsuperscript{2,3\textcurrency},
% Name4 Surname\textsuperscript{2},
% Name5 Surname\textsuperscript{2\ddag},
% Name6 Surname\textsuperscript{2\ddag},
% Name7 Surname\textsuperscript{1,2,3*},
% with the Lorem Ipsum Consortium\textsuperscript{\textpilcrow}
\\
\bigskip
% \textbf{1} Affiliation Dept/Program/Center, Institution Name, City, State, Country
% \\
% \textbf{2} Affiliation Dept/Program/Center, Institution Name, City, State, Country
% \\
% \textbf{3} Affiliation Dept/Program/Center, Institution Name, City, State, Country
\textbf{1} The Department of Mechanical and Intelligent Systems Engineering, Graduate School of Informatics and Engineering, The University of Electro-Communications, 1-5-1 Chofugaoka, Chofu, Tokyo, 182-8585 Japan

\textbf{2} Sony Corporation, Sony City Osaki, 2-10-1 Osaki, Shinagawa-ku, Tokyo, 141-8610 Japan
\\
\bigskip

% Insert additional author notes using the symbols described below. Insert symbol callouts after author names as necessary.
% 
% Remove or comment out the author notes below if they aren't used.
%
% Primary Equal Contribution Note
%\Yinyang These authors contributed equally to this work.

% Additional Equal Contribution Note
% Also use this double-dagger symbol for special authorship notes, such as senior authorship.
%\ddag These authors also contributed equally to this work.

%\textcurrency Current Address: Dept/Program/Center, Institution Name, City, State, Country % change symbol to "\textcurrency a" if more than one current address note
% \textcurrency b Insert second current address 
% \textcurrency c Insert third current address


% Group/Consortium Author Note
%\textpilcrow Membership list can be found in the Acknowledgments section.

% Use the asterisk to denote corresponding authorship and provide email address in note below.
%* correspondingauthor@institute.edu
* ynakata@uec.ac.jp

\end{flushleft}
\section*{Abstract}
Robots designed for therapy and social interaction are often intended to evoke a sense of ``animacy'' in humans. 
% Robots designed for therapy and social interaction are often intended to evoke a sense of ``animacy'' in humans. 
While many robots have been developed to imitate life-like appearance and joint movements, relatively little attention has been given to the effect of whole-body expansion and contraction---volume-changing movements commonly observed in living organisms---on perceived animacy. 
% While many robots have been developed to imitate appearance and joint movements, little attention has been paid to the effect of whole-body expansion and contraction---volume-changing movements observed in living organisms---on the perception of animacy.
In this study, we developed MOFU (MOrphing Fluffy Unit), a mobile robot capable of whole-body expansion and contraction using a single motor enclosed within a fluffy exterior. 
% In this study, we designed and developed a mobile robot called ``MOFU (Morphing Fluffy Unit)'', which can perform whole-body expansion and contraction with a single motor and is covered with a fluffy exterior. 
MOFU employs a ``Jitterbug'' structure, a geometric transformation mechanism that enables smooth volume changes in diameter from approximately 210 mm to 280 mm with a single actuator. 
% MOFU adopts a ``Jitterbug'' structure, a geometric transformation mechanism that enables smooth volume change in diameter from approximately 210 mm to 280 mm using a single actuator. 
Additionally, it is equipped with a differential two-wheel drive mechanism for locomotion.
% In addition, it is equipped with a differential two-wheel drive mechanism for locomotion.
We conducted an online survey using videos of MOFU's behavior to evaluate the effect of expansion--contraction movements on animacy perception. 
% To evaluate the effect of expansion-contraction movements on animacy perception, we conducted an online survey using videos of MOFU's behavior.
Participants rated their impressions using the Godspeed Questionnaire Series. 
% Participants rated their impressions using the Godspeed Questionnaire Series. 
First, we compared participants’ ratings across videos of MOFU in stationary states, with and without expansion--contraction, and separately, with and without rotational motion. 
Both the expansion--contraction and rotational movements independently increased perceived animacy. 
% First, we compared videos of MOFU in a stationary state with and without expansion-contraction and, separately, with and without rotational motion, and found that each movement independently increased perceived animacy. 
Second, we hypothesized that presenting two MOFUs simultaneously would enhance perceived animacy further; however, this prediction was not supported, as no significant difference was observed. 
% Second, we hypothesized that presenting two MOFUs simultaneously would increase perceived animacy compared with a single presentation; however, this prediction was not supported, as no significant difference was observed. 
Exploratory analyses were also performed across the four dual-robot motion conditions. 
% In addition, we conducted exploratory analyses comparing the four dual-robot motion conditions. 
Third, when expansion--contraction was combined with locomotion, animacy ratings were higher than for locomotion alone.
% Third, when expansion-contraction was combined with locomotion, the animacy ratings were higher than locomotion alone.
These results suggest that volume-changing movements, such as expansion and contraction, enhance the perception of animacy in robots. 
Consequently, incorporating physical volume change may be an important design element in the development of future robots aimed at shaping human impressions.
\section*{Introduction}

Robots designed for therapy and social interaction are often intentionally created to evoke a sense of ``animacy'' in users. 
% Research and development of robots designed for therapy and social interaction have been actively pursued.
For instance, the seal-shaped robot PARO has been employed to promote emotional engagement among elderly individuals, particularly those with dementia, and its therapeutic effects have been reported \cite{Shibata2012-qq}. 
% For example, the seal-shaped robot PARO has been employed as a tool to promote emotional engagement with elderly individuals, particularly those with dementia, and its therapeutic effects have been reported \cite{Shibata2012-qq}.
Similarly, the bear-shaped robot Huggable has been shown to reduce anxiety and stress in hospitalized children through interactive engagement \cite{Jeong2017-bv}. 
% Similarly, the bear-shaped robot Huggable has been suggested to reduce anxiety and stress in hospitalized children through interactive engagement \cite{Jeong2017-bv}.
These robots typically integrate elements such as animal-like appearances, soft tactile surfaces, and responsive behaviors, all of which contribute to shaping human--robot relationships.
% These robots typically combine elements such as animal-like appearance, soft tactile surfaces, and responsive behaviors, which are considered to play a role in shaping human--robot relationships.
% セラピーやソーシャルインタラクションを目的としたロボットの開発が進められている。その多くでは、ユーザがロボットに対してある種の「生き物らしさ（アニマシー）」を感じるようなデザインが意図されている。たとえば、アザラシ型ロボットのPAROは、高齢者、特に認知症患者との情緒的交流を促す手段として用いられ、その効果が報告されている\cite{Shibata2012-qq}。また、クマ型ロボットのHuggableは、小児患者とのインタラクションを通じて、入院中の不安やストレスの軽減に寄与する可能性が示されている\cite{Jeong2017-bv}。こうしたロボットでは、動物的な外観や柔らかい触感、反応的な動作といった要素が組み合わされており、そうした要素が人との関係性の形成に影響を与えることが示唆されている。

\begin{figure}[htbp]
  \centering
  \includegraphics[width=0.8\textwidth]{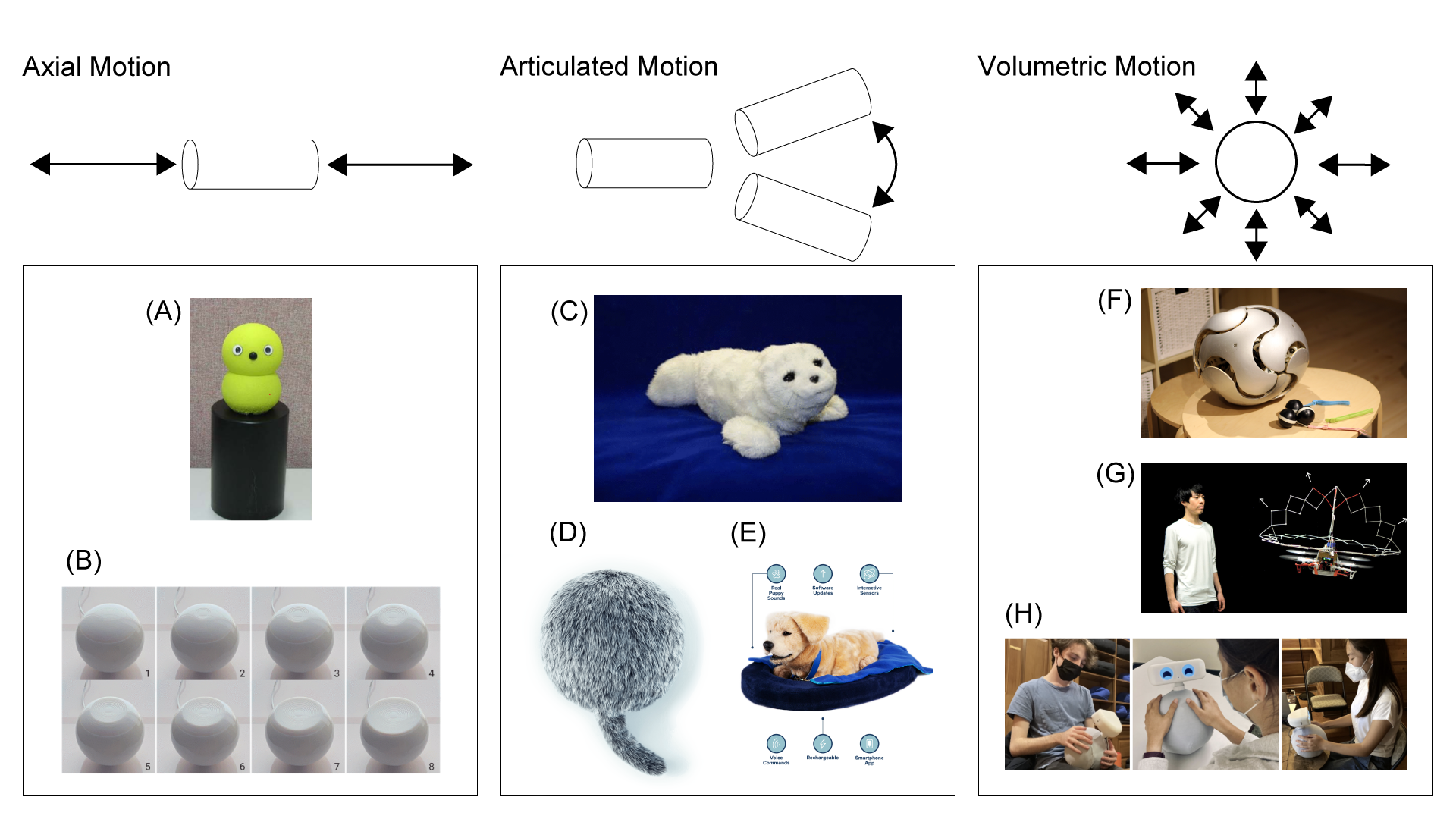}
  \caption{Classification of robot body motions into three categories---axial, articulated, and volumetric motion—--based on structural characteristics of movement.}
  \label{robotByMoves}
\end{figure}

Wolf et al. demonstrated that the perception of animacy in humans, animals, and machines is influenced not only by appearance but also by the presence or absence of movement \cite{Wolf2020-ey}. 
% Wolf et al. demonstrated that the perception of animacy in humans, animals, and machines is significantly influenced not only by their appearance but also by the presence or absence of movement \cite{Wolf2020-ey}.
This finding indicates that animacy in robots is shaped not solely by external appearance but also by the nature of their movements and responses. 
% This finding suggests that animacy in robots is shaped not solely by external appearance but also by the nature of their movements and responses. 
Lifelike movements, in particular, are considered essential for robots to be perceived as intuitive and approachable. 
% In particular, lifelike movements are considered critical for robots to be perceived as more intuitive and approachable. 
Previous studies have shown that both a robot's appearance and the characteristics of its movements strongly affect impressions of animacy, likability, trustworthiness, and unpleasantness \cite{Castro-Gonzalez2016-uf,10.1145/3344286}. 
% Indeed, it has been shown that not only a robot's appearance but also the characteristics of its movements strongly affect impressions of animacy, likability, trustworthiness and unpleasantness \cite{Castro-Gonzalez2016-uf,10.1145/3344286}.
To organize the range of robot movements explored in human--robot interaction research, we classify them into three categories—--axial, articulated, and volumetric motion—--based on prior literature and our own analysis (Fig.~\ref{robotByMoves}).
% To organize the variety of robot movements explored for human--robot interaction, we classify them into three categories—axial motion, articulated motion, and volumetric motion—based on previous literature and our own synthesis (Fig.~\ref{robotByMoves}).

Axial motion encompasses linear extension and contraction, as observed in organisms such as earthworms and inchworms \cite{das2023earthworm}, as well as temporary retraction along the body axis, exemplified by Keepon \cite{kozima2009keepon}. 
% Axial motion includes linear extension and contraction, such as those observed in earthworms and inchworms \cite{das2023earthworm}, as well as temporary retraction along the body axis, as seen in Keepon \cite{kozima2009keepon}.
Robots that replicate the movements of worms or inchworms have typically been developed for bio-inspired purposes, such as locomotion in confined environments or adaptation to unstructured terrain. 
% Robots that mimic the movements of worms or inchworms have typically been developed for bio-inspired applications such as locomotion in confined spaces or environmental adaptability, and their effects on animacy perception have been only minimally evaluated.
Additionally, studies have explored smooth, unidirectional volumetric deformation to evoke lifelike impressions\cite{Iizawa2022-cn}. 
% Attempts have also been made to give a lifelike impression by using smooth one-directional volumetric deformation\cite{Iizawa2022-cn}.
However, the influence of these movements on the perceived animacy has been only minimally examined. 
Articulated motion involves the flexion and extension of joints, as well as the movements of continuous multi-joint structures, such as a backbone or flexible body parts, such as a tail \cite{Pleo10.1145/3610978.3640730,Qoobo10.1145/1810543.1810549}. 
% Articulated motion encompasses joint flexion and extension, as well as the movements of continuous multi-joint structures, such as a backbone, or flexible body parts, such as a tail \cite{Pleo10.1145/3610978.3640730,Qoobo10.1145/1810543.1810549}. 
% These segmented and localized movements have been employed to convey animacy and to express affective states.

In contrast, volumetric motion involves changes in the overall body volume through expansion and contraction. 
% In contrast, volumetric motion refers to a form of movement in which the overall body volume changes through expansion and contraction. 
In living organisms, this phenomenon occurs not only in physiological functions such as breathing but also in social displays, including courtship displays of peacocks and the throat pouch inflation of frigatebirds. 
% In living organisms, this is observed not only in physiological functions such as breathing but also in social display behaviors, such as peacock courtship displays or the throat pouch inflation of frigatebirds.
Articulated motion, in particular segmented and localized movements, has been used to convey animacy and express affective states. 
By incorporating volumetric motion, robots can achieve global and continuous shape changes that are challenging to realize with conventional joint-driven mechanisms. 
% Such movements enable robots to achieve global and continuous shape changes that are difficult to realize with conventional joint-driven mechanisms. 
Hence, volumetric motion has the potential to significantly expand the expressive capabilities of robots.

Building on this concept, several devices have been developed that incorporate expansion--contraction movements inspired by physiological rhythms, such as breathing or heartbeat. 
% Devices that incorporate expansion-contraction movements inspired by physiological rhythms such as breathing or heartbeat
For instance, Ommie \cite{Matheus2022-ah} is designed to mimic breathing motions with the goal of reducing users’ anxiety and tension.
% For example, Ommie \cite{Matheus2022-ah} is designed to mimic breathing motions with the aim of alleviating users' anxiety and tension. 
These devices typically employ localized volume changes in specific parts of the body to elicit emotional effects, such as fostering comfort and empathy. 
% Such devices typically employ localized volume changes in parts of the body to elicit emotional effects, such as promoting comfort and empathy.
In contrast, PufferBot \cite{Hedayati2020-mu} uses a whole-body expansion--contraction mechanism to support functions such as maintaining safe interpersonal distance or facilitating locomotion. 
% In contrast, PufferBot \cite{Hedayati2020-mu} employs a whole-body expansion-contraction mechanism to support functions such as maintaining safe interpersonal distance or enabling locomotion. 
However, in these cases, volumetric changes are primarily functional or behavioural; the potential of such movements to evoke animacy in humans has not been sufficiently explored.
% However, in these cases, volumetric changes are primarily utilized for functional or behavioral purposes, and the potential of such movements to evoke animacy in humans has not been sufficiently examined.

% 呼吸や心拍など、生理的なリズムに着想を得た膨張・収縮動作が導入されたデバイスもいくつか存在している。たとえば、Ommie \cite{Matheus2022-ah} は呼吸動作を模倣し、ユーザーの緊張や不安を緩和することを目的とした設計がなされている。このような例は身体の一部における小さな体積変化を通じて、安心感や共感の促進といった情緒的な効果を意図している。一方で、PufferBot \cite{Hedayati2020-mu} は身体全体の膨張・収縮機構を用いて、人の安全な距離の確保や移動といった機能を実現する試みがなされている。しかし、このような研究では体積変化は主に機能的・行動的な目的のために用いられており、その動作が人間に「生き物らしさ（アニマシー）」を喚起する可能性については十分に検討されていない。

This study aims to clarify the effect of whole-body expansion--contraction movements (volumetric motion) on the perception of animacy in robots. 
% The purpose of this study is to clarify how whole-body expansion-contraction movements (volumetric motion) influence the perception of animacy in robots. 
To this end, we designed and developed MOFU (MOrphing Fluffy Unit), a mobile robot with a ``Jitterbug'' structure, a geometric transformation mechanism that enables smooth volumetric change using a single actuator. 
% To this end, we designed and developed a mobile robot named MOFU (Morphing Fluffy Unit), which adopts a ``Jitterbug'' structure, a geometric transformation mechanism that enables smooth volumetric change using a single actuator. 
MOFU uses a single motor-driven linear mechanism for expansion and contraction and a differential two-wheel drive for locomotion, all covered with a soft, fluffy exterior. 
% MOFU expands and contracts through a single motor-driven linear mechanism, is equipped with a differential two-wheel drive for locomotion, and is covered with a soft, fluffy exterior. 
A mathematical model was used to predict dimensional changes during expansion and contraction; experimental results confirmed that the robot operated as designed. 
% In addition, a mathematical model was used to predict dimensional changes during expansion and contraction, and experimental results confirmed that the robot operated as designed. 
To evaluate MOFU’s behavior, we conducted an online survey. 
Participants watched videos of MOFU and rated their impressions using the Animacy subscale of the Godspeed Questionnaire Series. 
% For evaluation, we conducted an online survey in which participants viewed videos of MOFU's behavior and rated their impressions using the Animacy subscale of the Godspeed Questionnaire Series. 
Specifically, we examined three conditions: (i) effects of expansion--contraction and rotational motions in a stationary state (single-robot condition), (ii) impressions when two MOFUs were presented simultaneously (dual-robot condition), and (iii) effect of combining expansion--contraction with locomotion (locomotion condition). 
% Specifically, we examined three conditions: (i) the effects of expansion-contraction and rotational motions in a stationary state (single-robot condition), (ii) impressions when two MOFUs were presented simultaneously (dual-robot condition), and (iii) the effect of combining expansion-contraction with locomotion (locomotion condition).     
The dual-robot condition included four motion variants analogous to those in the single-robot condition to examine generalization, with comparisons among these dual-robot variants treated as exploratory.
% In the dual-robot condition, the four motion variants analogous to the single-robot condition were included to examine generalization; comparisons among these dual-robot variants were treated as exploratory.

% 本研究の目的は、全身的な膨張収縮動作（volumetric motion）がロボットのアニマシー知覚に与える影響を明らかにすることである。 そのために、可変面構造（Jitterbug構造）を採用した移動型ロボット「MOFU（Morphing Fluffy Unit）」を設計・開発した。MOFUは、単一のモータ駆動の直動機構によって滑らかに膨張・収縮し、作動二輪による移動機構と柔らかな起毛素材の外装を備えている。さらに、数学モデルに基づき膨張収縮時の寸法変化を予測し、設計通りの動作が実現されていることを実験により確認した。評価実験では、MOFUの動作を撮影した複数の動画を用い、Godspeed Questionnaire SeriesのAnimacy尺度に基づくオンラインアンケート調査を実施した。具体的には、①静止状態における膨張収縮動作と旋回動作の有無を操作してそれぞれの主効果を検証し、②移動中に膨張収縮を組み合わせた場合のアニマシー評価、③2台のMOFUによる同時動作による印象の変化について検討した。

The contributions of this study are as follows:
\begin{enumerate}
\item \textbf{Development of a mobile robot capable of volumetric expansion and contraction:} We designed and implemented MOFU, a mobile robot that employs a Jitterbug structure to effortlessly expand and contract its body diameter from approximately 210 mm to 280 mm using a single motor. A mathematical model was employed to predict dimensional changes. Experiments successfully validated the robot’s performance, confirming its functionality as intended. % We designed and implemented MOFU, a mobile robot that utilizes a Jitterbug structure to achieve smooth expansion and contraction of its body diameter from approximately 210 mm to 280 mm with a single motor. A mathematical model was used to predict dimensional changes, and experiments confirmed that the robot performed as designed.
\item \textbf{Empirical demonstration in a single-robot condition:} By comparing conditions with and without expansion--contraction and rotational motions when stationary (i.e., without locomotion), we demonstrated that expansion--contraction significantly enhances perceived animacy. % By comparing conditions with and without expansion-contraction and rotational motions when stationary (i.e., without locomotion), we showed that expansion-contraction significantly increases perceived animacy.
\item \textbf{Evaluation in a dual-robot condition:} We investigated the impressions of two robots presented simultaneously; the results did not indicate an increase in animacy compared to that of a single robot presentation. The within-dual-robot comparisons among the four motion conditions were exploratory. % We examined impressions when two robots were presented simultaneously, but the results did not show increased animacy compared with a single presentation. The within dual-robot comparisons among the four motion conditions were exploratory.
\item \textbf{Evaluation in a locomotion condition:} We demonstrated that when combined with locomotion, expansion--contraction increased perceived animacy compared to locomotion alone. % We demonstrated that expansion-contraction increased perceived animacy when combined with locomotion compared with locomotion alone.
\end{enumerate}

% 本研究の貢献は以下通りである．
% \begin{enumerate}
%     \item 膨張収縮動作を実現する移動型ロボット「MOFU」の開発：Jitterbug構造を活用し、単一のモータで直径約140 mmから230 mmまで滑らかに膨張・収縮可能な移動型ロボットを設計・実装した。設計には寸法変化の数学モデルを用い、実験により設計通りの動作を確認した。
%     \item 膨張収縮動作がアニマシー評価に与える影響の実証：静止状態において膨張収縮および旋回の有無を操作した条件を比較し、膨張収縮動作がアニマシーを有意に高める主要因であることを明らかにした。
%     \item 複数条件での追加的知見の提示：移動中の膨張収縮や、2台のロボットによる同時動作といった条件において、膨張収縮動作がアニマシー知覚にどのように影響を与えるかを評価した。
% \end{enumerate}

\newpage
\section*{Design and Development of MOFU}
This section describes the design and implementation of the developed robot, MOFU.

% This section describes the design and implementation of the developed robot, MOFU. 
We first outline the concept, followed by detailed explanations of the mechanical design, electrical and control systems, and the prototyping and implementation process. 
% First, the design concept is outlined, followed by explanations of the mechanical design, electrical and control systems, and the prototyping and implementation. 
Next, we present the modeling and validation of the expansion--contraction mechanism, and finally, we summarize the robot motions that can be achieved with MOFU.
% Next, the modeling and validation of the expansion-contraction mechanism are presented, and finally, the robot motions achievable with MOFU are summarized.

%本章では，開発したロボットMOFUの設計と実装について述べる．まず，設計思想を整理した上で，機械設計，電気・制御システム，試作と実装を順に説明する．続いて，膨張収縮機構のモデリングと検証について述べ，最後に，開発したMOFUが実現可能な動作について整理する．

\begin{figure}[htbp]
    \centering
    \includegraphics[width=\linewidth]{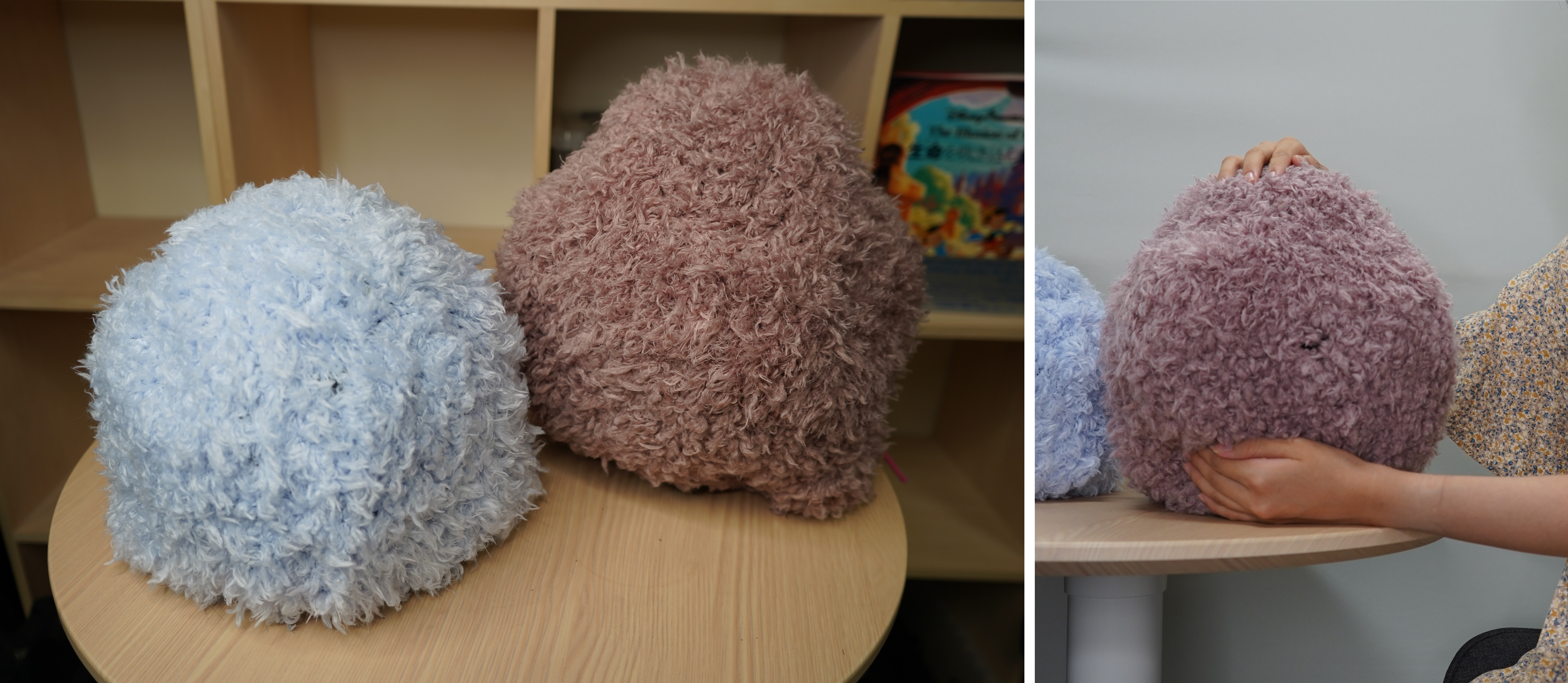}
    \caption{Two color variations of MOFU and a MOFU being gently stroked. MOFU was designed to evoke a sense of life-likeness through its volumetrically deformable body.}
    \label{fig: mofu_strike_img}
\end{figure}

\subsection*{Design concept}
The robot developed in this study was designed around an expansion--contraction mechanism to evoke a sense of animacy, complemented by several additional features. 
% The robot developed in this study was designed with an expansion-contraction mechanism at its core to realize a sense of animacy, supplemented by several additional features. 
Here, we describe the specific design concepts of quiet operation, mobility, and a fluffy exterior.
% Here, we describe the specific concepts of quiet operation, mobility, and a fluffy exterior.

As animal muscles typically do not produce sound during movement, minimizing operational noise was considered essential for achieving a sense of ``animacy.'' 
% Since animal muscles do not produce sound during movement, minimizing operational noise was considered an essential factor in achieving ``animacy.''
To this end, the robot uses direct-drive motors without gears and incorporates components with low sliding noise. This design suppresses mechanical sounds during motion, reducing the perception of mechanical artifacts among users.
% Therefore, the robot employs direct-drive motors without gears and components with low sliding noise to ensure quiet operation. 
% This design suppresses mechanical sounds during motion and reduces mechanical impressions perceived by users.
When measured in an indoor environment with ambient noise, the sound level during both locomotion and expansion--contraction remained below 50 dB, as recorded by a sound level meter (Testo 816-1) placed approximately 50 $cm$ from MOFU.

Animals that humans commonly interact with can move autonomously. 
% Animals that humans commonly interact with are capable of autonomous locomotion. 
Accordingly, the robot was equipped with a mobility mechanism. 
% Accordingly, the robot was equipped with a mobility mechanism. 
Structural simplicity and compactness was maintained by adopting wheeled locomotion. 
% To ensure structural simplicity and compactness, wheeled locomotion was adopted. 
A differential two-wheel drive mechanism enables both straight movement and rotation on a horizontal plane. 
% A differential two-wheel drive mechanism was implemented, enabling both straight movement and rotational motion on a horizontal plane.
Wheels were selected over legs to prioritize stability, as expansion--contraction was the primary interaction element. Additionally, the use of an internal power supply and elimination of external cables enhance the robot’s autonomous appearance.
% Wheels were chosen over legs to prioritize stability, as expansion-contraction was positioned as the primary interaction element.
% In addition, by incorporating an internal power supply and eliminating external cables, the robot emphasizes an autonomous appearance.

Furthermore, many animals are covered with skin or fur, which contributes to impressions of animacy. 
% Furthermore, many animals are covered with skin or fur, and such textures are considered to contribute to impressions of animacy.
Hence, the robot was enveloped in a soft, fluffy cover to emulate this characteristic. 
% To emulate this characteristic, the robot adopts a soft, fluffy cover. 
Previous studies comparing robots of the same design but with soft versus hard surface textures have shown that soft textures can reduce psychological and physiological stress, suggesting their potential effectiveness in robot therapy \cite{2019-uw}.
% Previous studies have compared robots with soft versus hard surface textures of the same design, and have shown that soft textures enhance the alleviation of psychological and physiological stress, suggesting their potential effectiveness in robot therapy \cite{2019-uw}.

\begin{figure}[htbp]
  \centering
  % 1st row --------------------------------
  \begin{subfigure}[b]{0.48\linewidth}
    \centering
    \includegraphics[width=\linewidth]{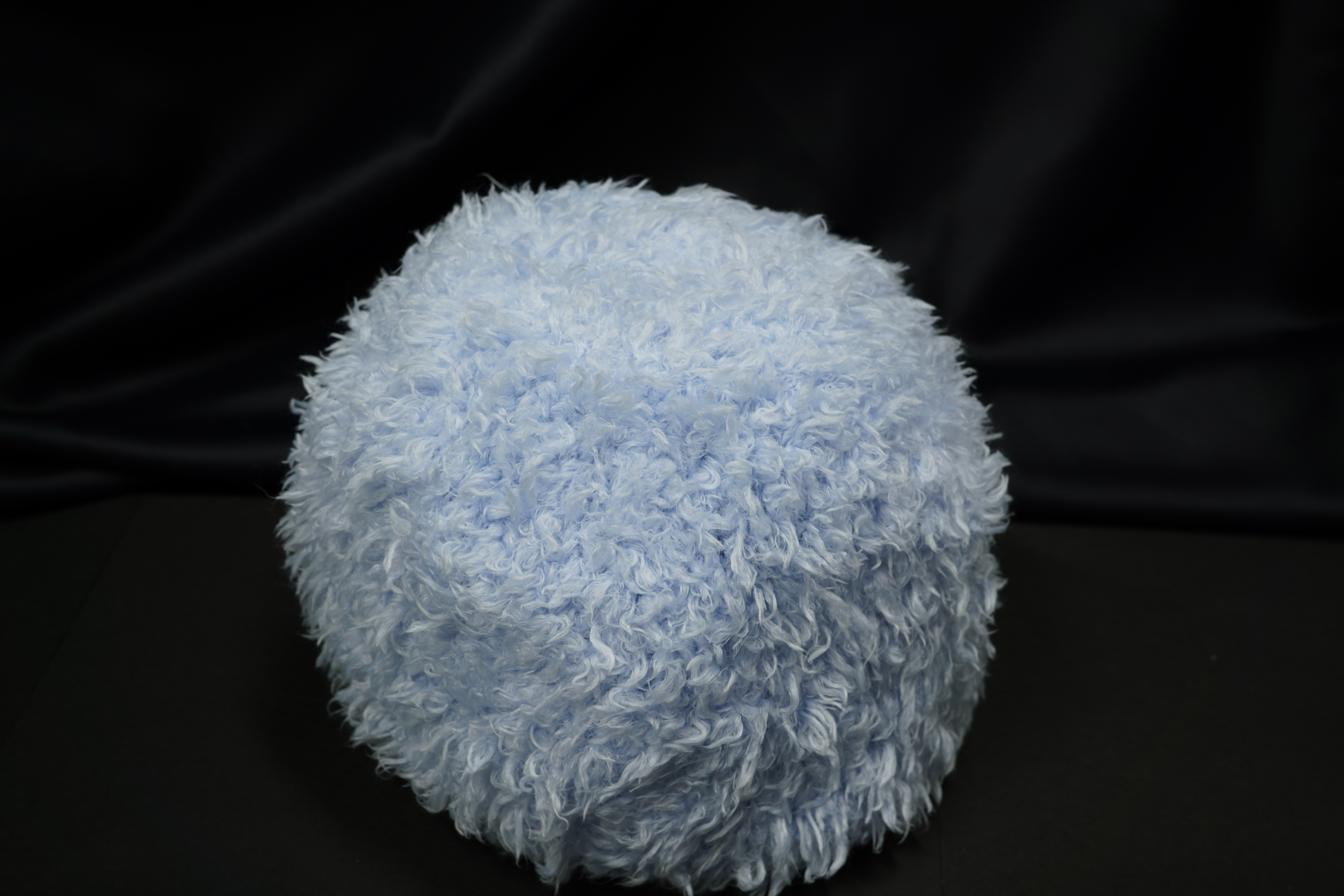}
    \caption{}
    \label{fig:mofu_c}
  \end{subfigure}\hfill
  \begin{subfigure}[b]{0.48\linewidth}
    \centering
    \includegraphics[width=\linewidth]{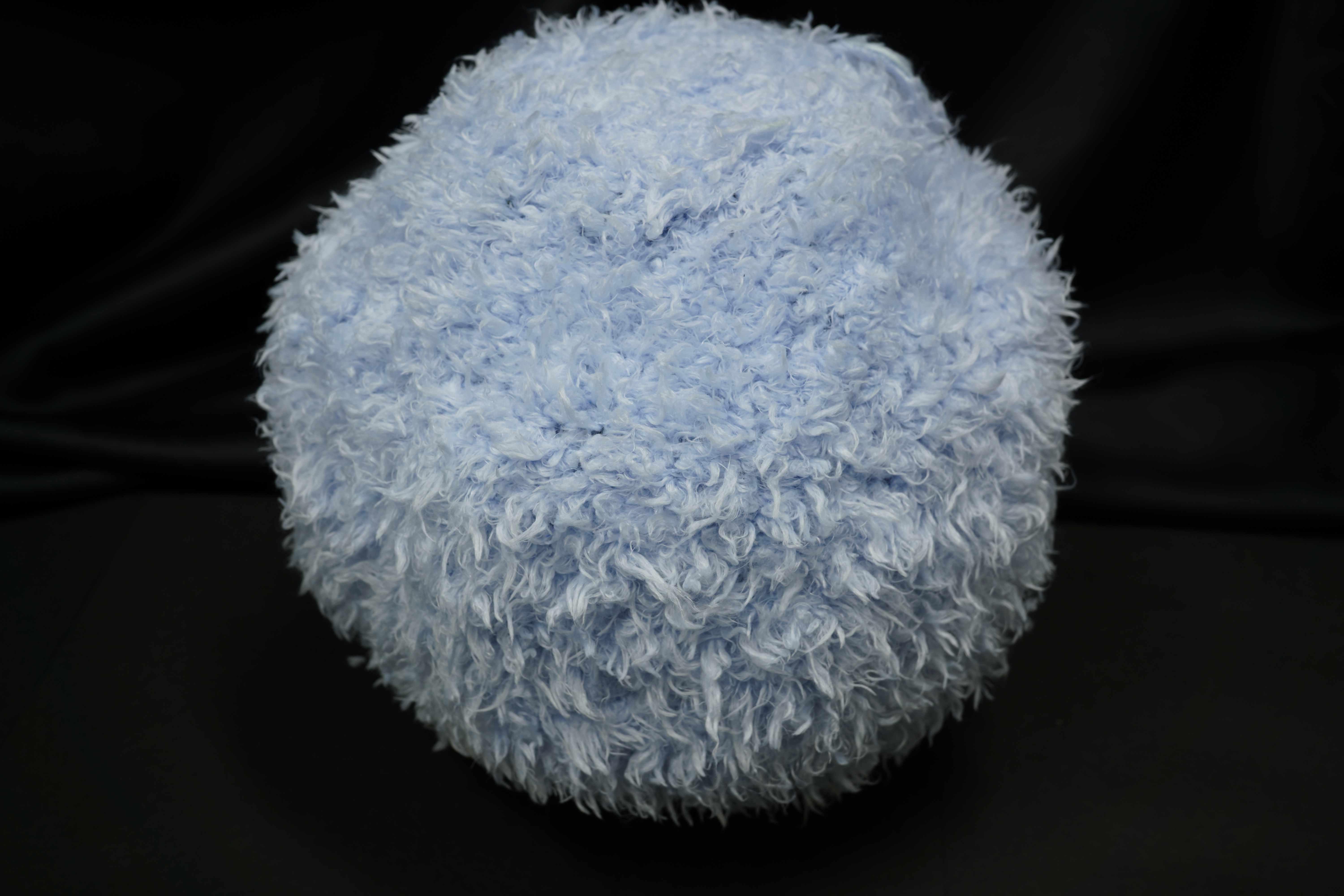}
    \caption{}
    \label{fig:mofu_e}
  \end{subfigure}

  \vspace{0.6em}

  % 2nd row --------------------------------
  \begin{subfigure}[b]{0.48\linewidth}
    \centering
    \includegraphics[width=\linewidth]{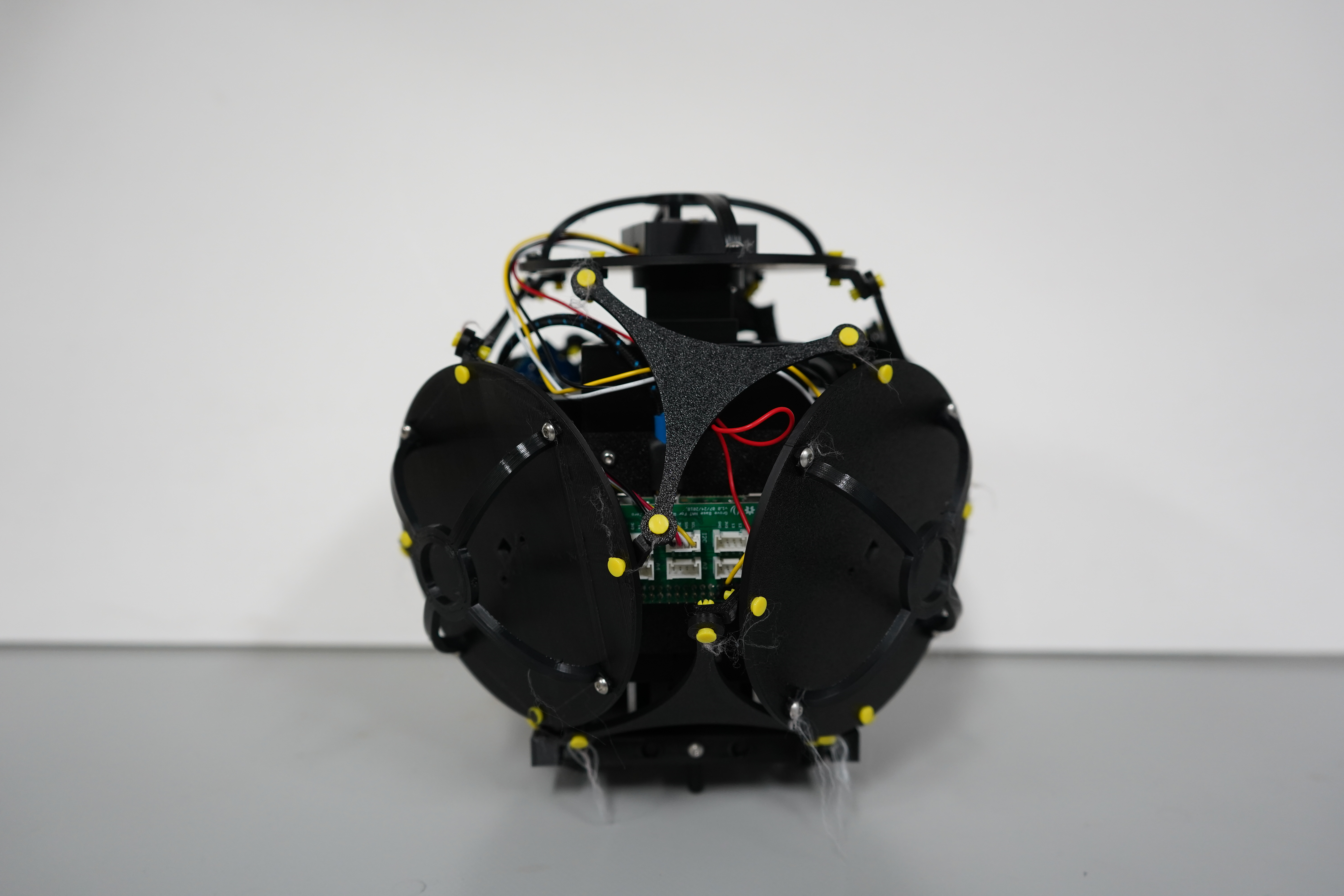}
    \caption{}
    \label{fig:mofu_bone_c}
  \end{subfigure}\hfill
  \begin{subfigure}[b]{0.48\linewidth}
    \centering
    \includegraphics[width=\linewidth]{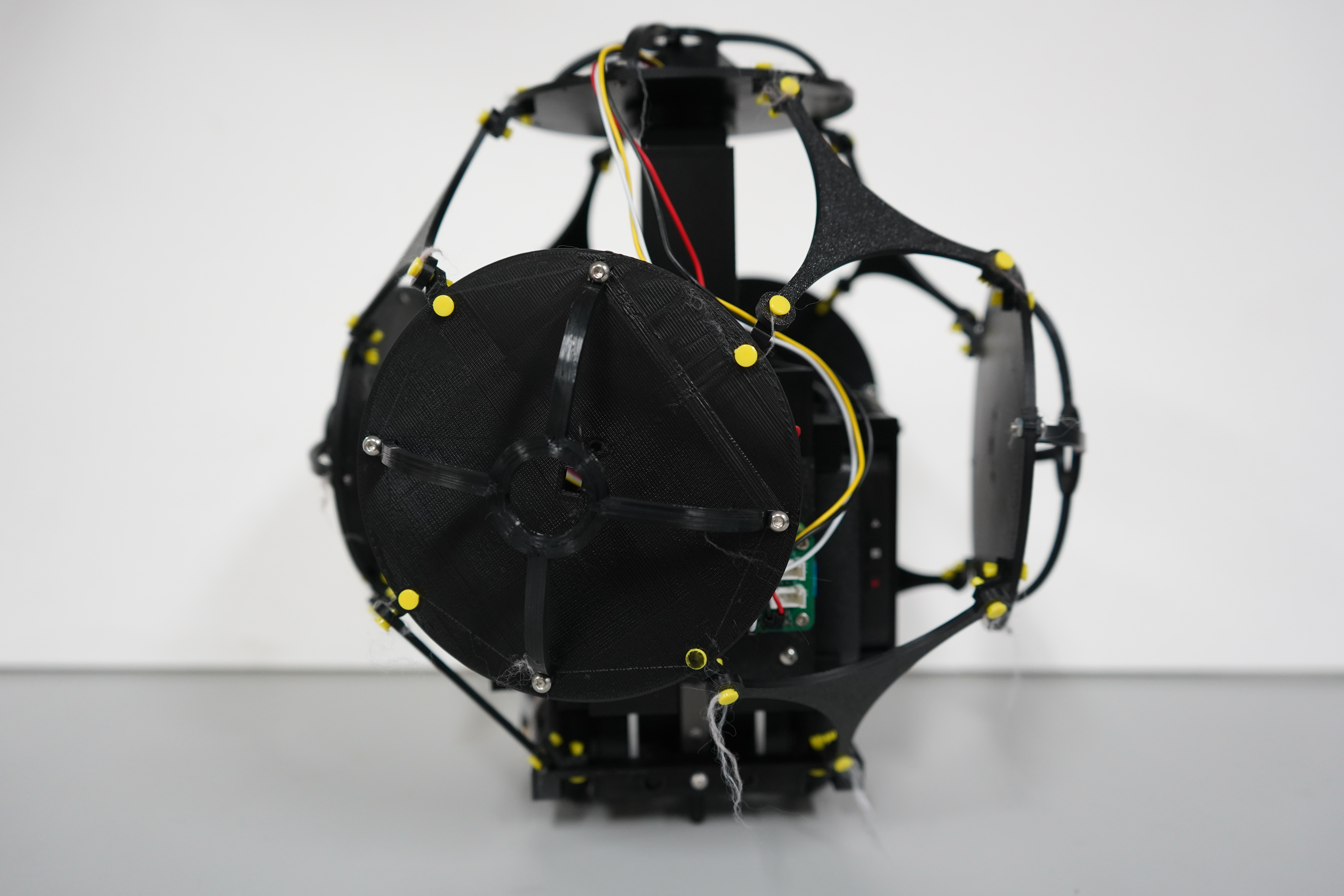}
    \caption{}
    \label{fig:mofu_bone_e}
  \end{subfigure}

  \vspace{0.6em}

  % 3rd row --------------------------------
  \begin{subfigure}[b]{0.48\linewidth}
    \centering
    \includegraphics[width=\linewidth]{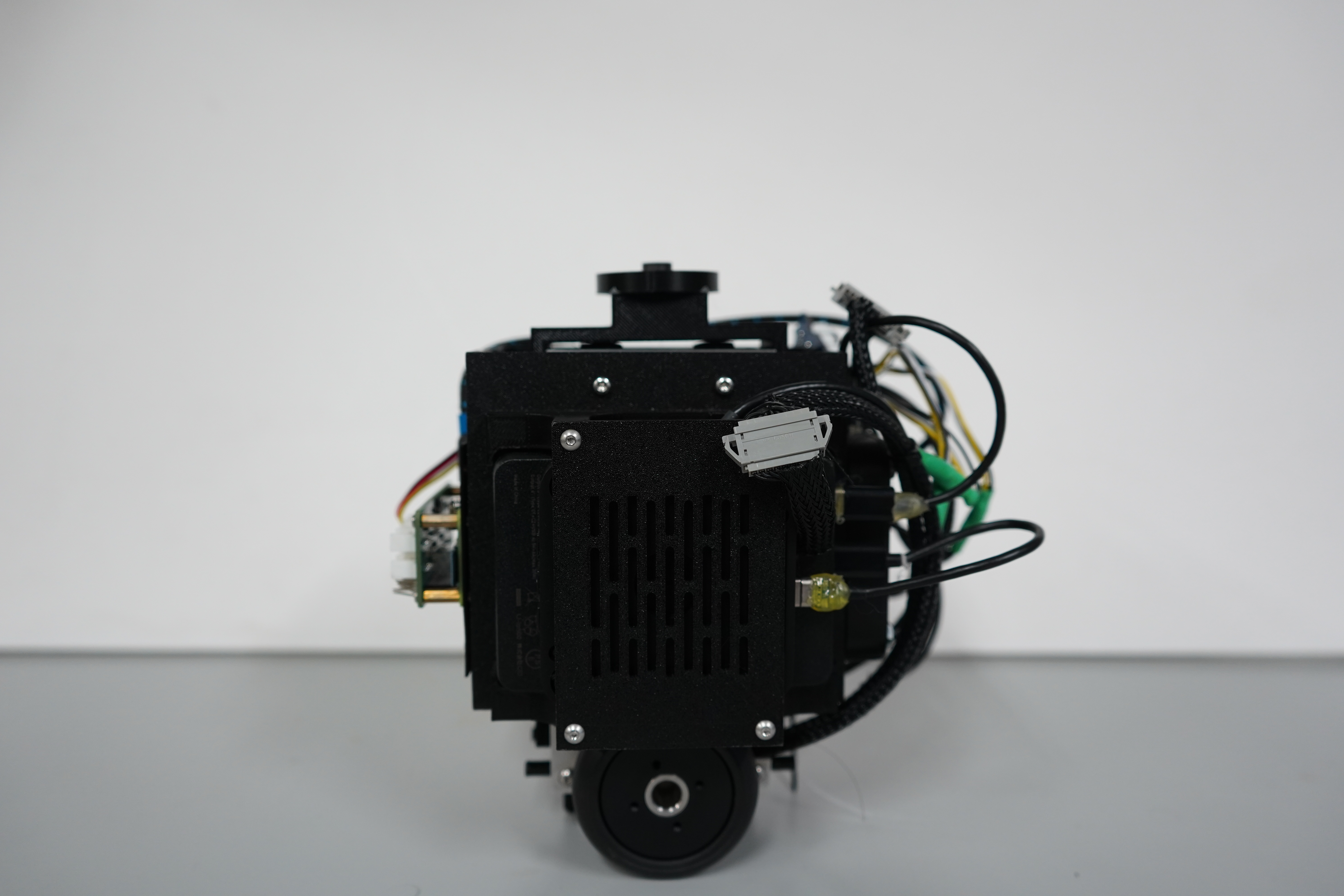}
    \caption{}
    \label{fig:mofu_mecha_c}
  \end{subfigure}\hfill
  \begin{subfigure}[b]{0.48\linewidth}
    \centering
    \includegraphics[width=\linewidth]{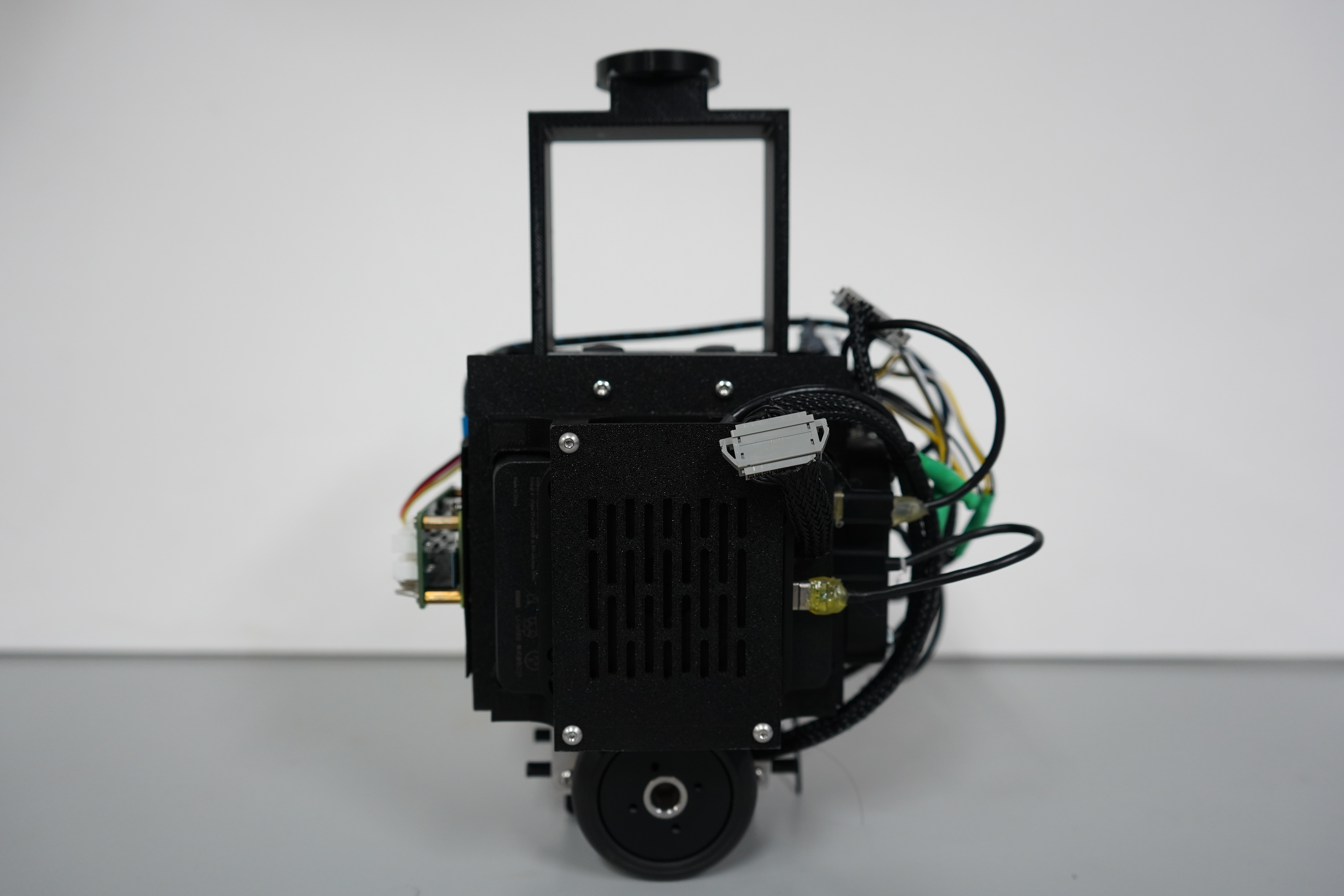}
    \caption{}
    \label{fig:mofu_mecha_e}
  \end{subfigure}

  \caption{Appearance of MOFU under different conditions: (a) with fluffy cover, contracted; (b) with fluffy cover, expanded; (c) without cover, contracted; (d) without cover, expanded; (e) without mechanism, contracted; (f) without mechanism, expanded.}
  \label{fig:mofu}
\end{figure}

\subsection*{Mechanical design}
MOFU has a spherical shape (Fig.~\ref{fig:mofu}) with a vertical height ranging from 210 to 280 mm, enabling whole-body expansion--contraction motion. 
% As shown in Fig.~\ref{fig:mofu}, MOFU has a spherical shape whose vertical height varies between 210 and 280 mm, enabling whole-body expansion-contraction motion.
This motion is achieved using a Jitterbug structure, where vertical displacement is controlled by a linear mechanism composed of a 20 mm lead screw (MTSPW1220, MISUMI Group Inc.) driven by a direct-drive motor (KM-1S-M4021, Keigan Inc.). 
% This motion is realized by a Jitterbug structure, in which the vertical displacement is controlled by a linear mechanism consisting of a 20 mm lead screw (MTSPW1220, MISUMI Group Inc.) driven by a direct-drive motor (KM-1S-M4021, Keigan Inc.). 
The Jitterbug structure, previously implemented in the modular robot AuxBot developed by Chin et al., enables expansion--contraction of the robot’s body using a single actuator \cite{Chin2023-yg}\cite{Lipton2019-mz}.
% The Jitterbug structure has also been employed in the modular robot AuxBot developed by Chin et al., and is characterized by achieving expansion-contraction with a single actuator \cite{Chin2023-yg}\cite{Lipton2019-mz}.

\begin{figure}[t]
    \centering
      % 1行目--------------------------------
      \begin{subfigure}[b]{0.48\linewidth}
        \centering
        \includegraphics[width=\linewidth]{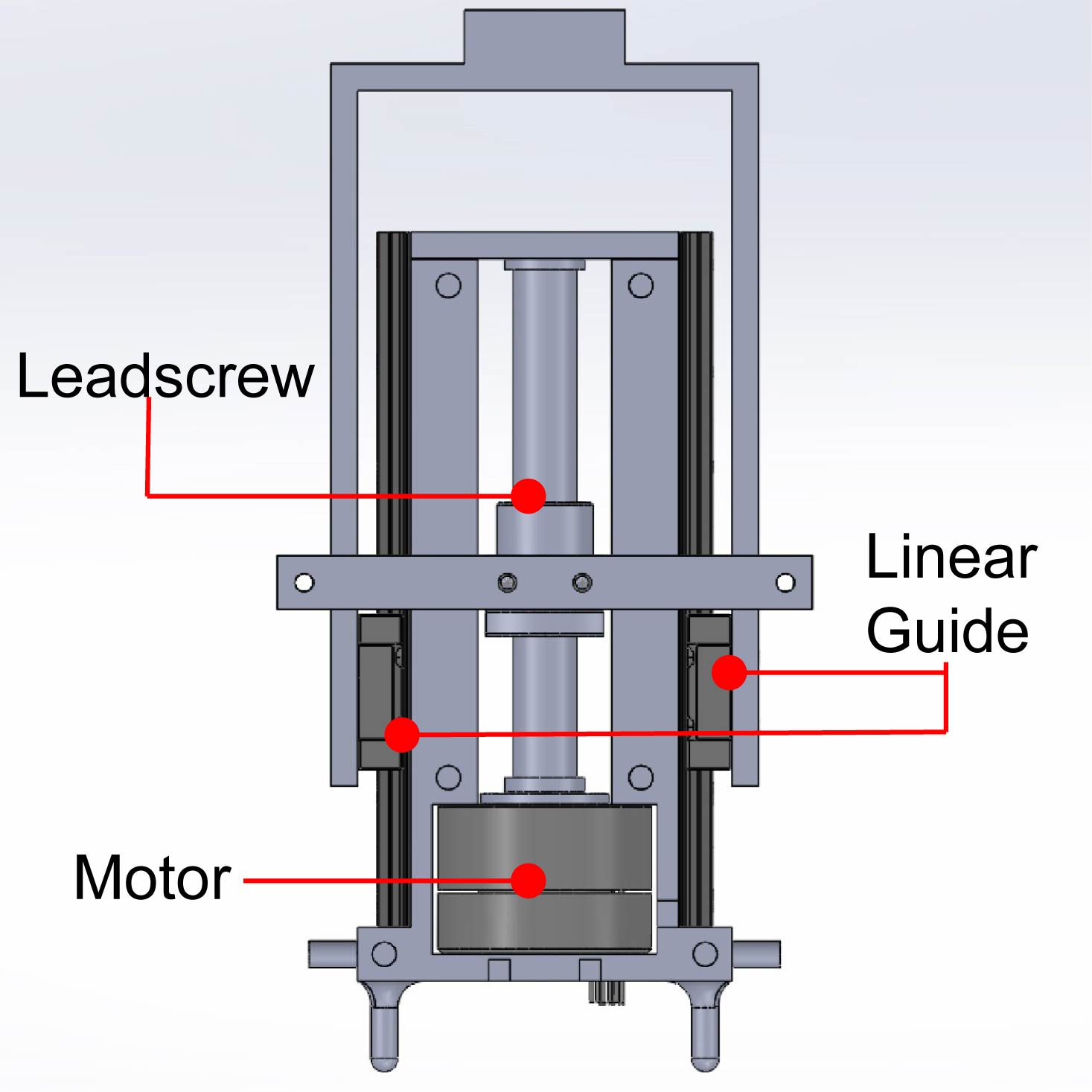}
        \subcaption{}\label{fig:screw_guide}
      \end{subfigure}\hfill
      \begin{subfigure}[b]{0.48\linewidth}
        \centering
        \includegraphics[width=\linewidth]{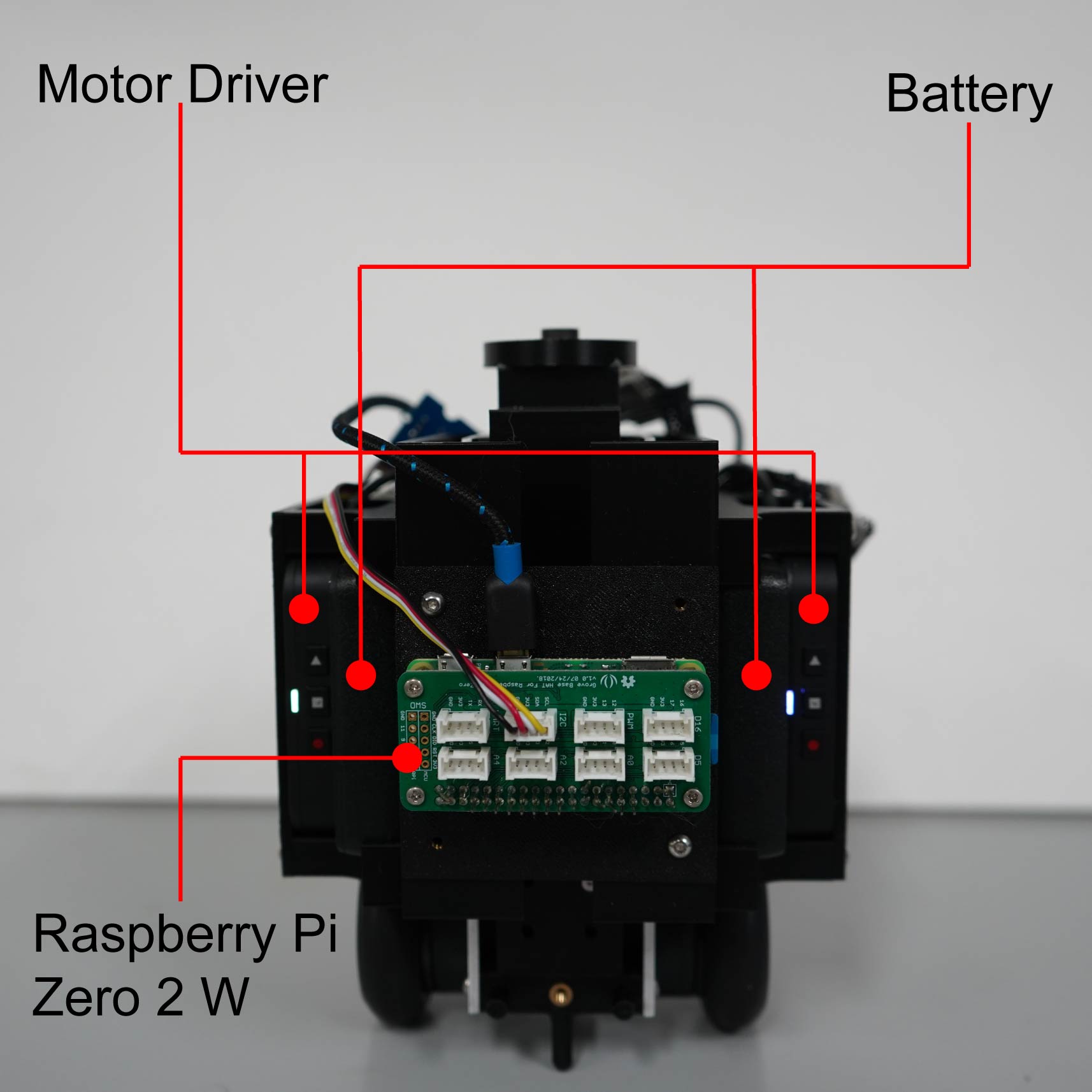}
        \subcaption{}\label{fig:battery_controller}
      \end{subfigure}
      \begin{subfigure}[c]{0.48\linewidth}
        \centering
        \includegraphics[width=\linewidth]{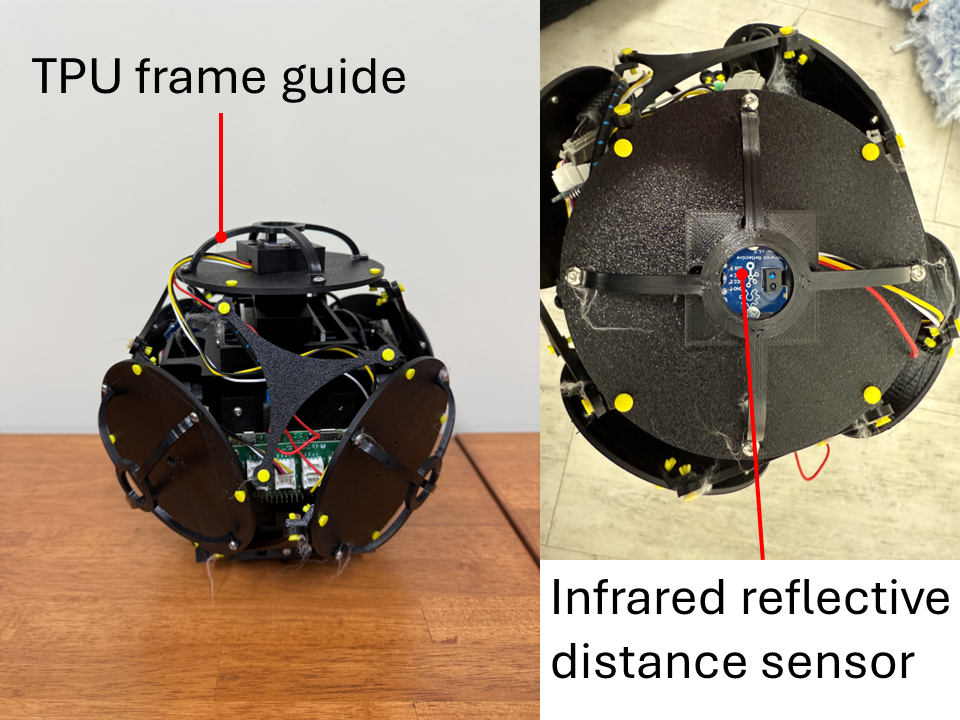}
        \subcaption{}\label{fig:touch_sensor}
      \end{subfigure}
    \caption{Mechanical design of MOFU:(a) Arrangement of the leadscrew and linear guides.
    (b) Arrangement of the motor driver, battery, and controller. 
    (c) Appearance of TPU frame guide and the mounting position of the touch sensor}
\end{figure}

The Jitterbug structure is inherently polyhedral; hence, 3-D printed TPU frames were attached to five faces, excluding the bottom face, to give the overall exterior a more spherical appearance. 
% Since the Jitterbug structure itself is polyhedral, 3-D printed TPU frames were attached to five faces excluding the bottom face, thereby making the overall exterior closer to a spherical form. 
Additionally, an infrared reflective distance sensor (Grove - Infrared Reflective Sensor, Seeed Studio) was mounted on the inner side of the top frame (Fig.~\ref{fig:touch_sensor}). 
% In addition, an infrared reflective distance sensor (Grove - Infrared Reflective Sensor, Seeed Studio) was installed on the inner side of the top frame (Fig.~\ref{fig:touch_sensor}). 
This sensor measures the distance to the fluffy cover to detect contact. 
% This sensor detects contact by measuring the distance to the fluffy cover. 
Although it was not used in the experiments reported here, it was implemented to enable potential future contact-based interactions. 
% Although it was not used in the experiments reported in this study, it was implemented with the aim of enabling future contact-based interaction. 
Similar sensors can be installed on other frames to easily expand the number of contact-detectable areas.
% Similar sensors can also be installed beneath the other frames, making it possible to easily extend the number of contact-detectable areas.

A quiet linear guide (TK-04, Igus K.K.) was adopted in combination with the direct-drive motor for the linear mechanism (Fig.~\ref{fig:screw_guide}) to ensure quiet operation and backdrivability, thereby minimizing operating noise.
% As shown in Fig.~\ref{fig:screw_guide}, in order to ensure quiet operation and backdrivability, a quiet linear guide (TK-04, igus K.K.) was adopted in combination with the direct-drive motor for the linear mechanism. 
% This minimized operating noise.
Locomotion on the horizontal plane was enabled by a differential two-wheel drive system, using two additional direct-drive motors of the same type to drive the wheels. 
% Furthermore, locomotion on the horizontal plane was made possible by constructing a differential two-wheel drive system, in which two additional direct-drive motors of the same type were used to drive the wheels. 
Rubber tires (NARROW TIRE SET, 58 mm DIA., TAMIYA, INC.) suitable for the motor size were employed. 
% Rubber tires (NARROW TIRE SET, 58 mm DIA., TAMIYA, INC.) appropriate to the motor size were employed.
Sliding contact parts were added at the front and rear of the bottom surface to prevent tipping, ensuring stability by maintaining four points of contact with the ground.

The arrangement of the individual components is illustrated in Fig.~\ref{fig:battery_controller}.
% To prevent the robot from toppling, sliding contact parts were placed at the front and rear of the bottom surface, ensuring stability by maintaining four points of ground contact. The arrangement of the individual components is shown in Fig.~\ref{fig:battery_controller}.

% 図\ref{fig:mofu}に示すように，MOFUは垂直方向の高さ210〜280 mmの範囲で変化する球体形状を有し，全身の膨張収縮動作が可能である．この動作はJitterbug構造によって実現され，リード20 mmのすべりねじ（MTSPW1220, MISUMI Group Inc.）をダイレクトドライブモータ（KM-1S-M4021, Keigan Inc.）で駆動する直動機構により垂直方向の変位を制御している．Jitterbug構造は，ChinらのモジュラーロボットAuxBotにも見られ，単一のアクチュエータで膨張収縮を可能にする点が特徴である\cite{Chin2023-yg}\cite{Lipton2019-mz}．

% Jitterbug構造自体は多面体であるため，底面を除く5つの面にTPU製のフレームを取り付け，全体の外形が球体に近づくようにした．さらに，上面のフレーム内側には赤外線反射型距離センサ（Grove - Infrared Reflective Sensor, Seeed Studio）を搭載した（図\ref{fig:touch_sensor}）．このセンサはFluffy coverとの距離を計測することで接触状態を検知できる．なお，本研究の実験では使用していないが，将来の接触インタラクションを想定して搭載したものである．また，同様のセンサは他の面のフレーム下にも設置可能であり，検知可能箇所を容易に拡張できる．

% 図\ref{fig:screw_guide}に示すように，静音性とバックドライブ性を確保するため，ダイレクトドライブモータの採用に加え，直動機構のガイドには静音リニアガイド（TK-04, igus K.K.）を用いた．これにより動作音を最小限に抑えている．

% さらに，同型のダイレクトドライブモータを2個で車輪を駆動する，差動二輪駆動の移動機構を構成することで，水平面上の移動が可能となっている．モータのサイズに合ったゴムタイヤ（NARROW TIRE SET (58mm DIA.), TAMIYA, INC.）を用いた．ロボットが転倒しないように，ロボット底面の前後に地面と接地する摺動部を設け安定性を確保した（つまり，地面には4点で設置している）．
% 使用した各部品の構成図を図\ref{fig:battery_controller}に示す．

\subsection*{Electrical and control systems}
\begin{figure}[htbp]
    \centering
        \includegraphics[width=0.7\linewidth]{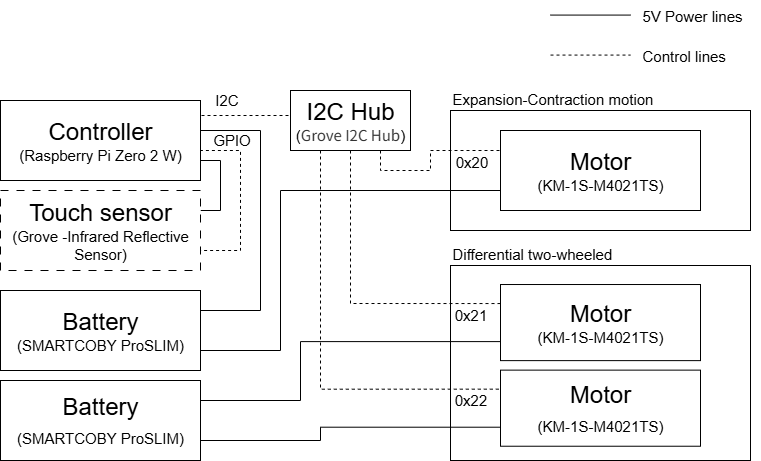}
    \caption{System configuration of MOFU}
    \label{fig:systemdiagram}
\end{figure}

The system configuration of MOFU is illustrated in Fig.~\ref{fig:systemdiagram}. 
% The system configuration of MOFU is shown in Fig.~\ref{fig:systemdiagram}. 
MOFU is powered by three direct-drive motors (KM-1S-M4021TS, Keigan Inc.), which are controlled via feedback by motor drivers according to position commands provided by a controller (Raspberry Pi Zero 2 W, Raspberry Pi Ltd). 
% MOFU operates with three direct-drive motors (KM-1S-M4021TS, Keigan Inc.), which are controlled via feedback by motor drivers according to the position commands provided by a controller (Raspberry Pi Zero 2 W, Raspberry Pi Ltd).
The PID control gains were set to $K_p = 5.0 , K_i = 10.0 , K_d = 0$. Power for both the motor drivers and the controller is supplied by a lithium-ion battery (SMARTCOBY Pro SLIM, CIO Co., Ltd.) that provides a shared 5 V supply.
% The PID control gains were set to $K_p = 5.0 , K_i = 10.0 , K_d = 0$. 
% Power for both the motor drivers and the controller is supplied by a lithium-ion battery (SMARTCOBY Pro SLIM, CIO Co., Ltd.) that provides a shared 5 V supply.
As mentioned earlier, an infrared reflective distance sensor is installed on the top surface of MOFU, although it was not used in the experiments reported in this study.
% Although not used in the experiments reported in this study, an infrared reflective distance sensor is installed on the top surface of MOFU. 
This sensor provides a binary signal indicating whether the top surface is in contact, which the controller can use to trigger motions or other interactions.
% This sensor provides a binary signal corresponding to the contact state of the top surface, which can be transmitted to the controller and used as a trigger for motion or other interactions.

MOFU can also detect collisions and perform responsive actions to contact with humans or walls by utilizing the wheel torque sensor and setting an appropriate threshold ($0.08$\si{\newton\meter}).
% MOFU can also detect collisions and perform responsive actions to contact with humans or walls by utilizing the wheel torque sensor and setting an appropriate threshold ($0.08$\si{\newton\metre}).

% MOFUのシステム構成を図\ref{fig:systemdiagram}に示す．開発したMOFUは3つのダイレクトドライブモータ（KM-1S-M4021TS, Keigan Inc.）をコントローラ（Raspberry Pi Zero 2 W, Raspberry Pi Ltd）から与えられた位置の指示値に応じてモータドライバのフィードバック制御により動作を行う．PID制御のゲインは$K_p=, K_i=, K_d=$とした．
% モータドライバとコントローラへは，リチウムイオンバッテリー（SMARTCOBY Pro SLIM, CIO co., ltd.）から供給される5V電源を共有する．
% 本研究の実験では使用しないが，MOFUの上面に搭載されている赤外線反射型距離センサによりMOFUの上面の接触状態を2値で取得できる．この情報は，コントローラへと送信して動作のトリガなどとして使用できる．
% 他にも，車輪のトルクセンサを用いて適切なしきい値（0.08）を設定することで，MOFUの衝突を検知し人や壁との接触に対する反応動作を行うことができる．

\subsection*{Prototyping and implementation}
The main specifications of MOFU are listed in Table~\ref{tab:spec}. 
% The main specifications of MOFU are shown in Table~\ref{tab:spec}. 
Key parameters, including the vertical height range, expansion--contraction speed, continuous operating time, and total weight, were measured and confirmed. 
% Parameters such as the range of vertical variation, expansion-contraction speed, continuous operating time, and total weight were measured and confirmed. 
These specifications were used to assemble the prototype by integrating the mechanical, electrical, and control systems. 
% Based on these specifications, the prototype was implemented by integrating the mechanical, electrical, and control systems. 
Additionally, the wheels were wrapped with yarn (SKI YARN Tenshi no Fur, Motohiro \& Co., Ltd.) to match the fluffy exterior and further reduce mechanical impressions.
% In addition, the wheels were covered with yarn (SKI YARN Tenshi no Fur, Motohiro \& Co., Ltd.) to provide consistency with the fluffy exterior and further reduce the mechanical impression.

\begin{table}[htbp]
\centering
\caption{Main specifications of MOFU}
\label{tab:spec}
\resizebox{\textwidth}{!}{%
\begin{tabular}{lll}
\toprule
Item & Value & Measurement condition/method \\
\midrule
Overall height (including wheels and frame) & 210--280 mm & Measured, maximum vertical point at rest \\
Distance between wheel contact points & 90 mm & Distance between ground contact points of the wheels \\
Wheel diameter & 58 mm & Rubber wheel diameter \\
Total battery capacity & 20000 mAh & -- \\
Total weight & 2091 g & Including battery and exterior cover \\
\bottomrule
\end{tabular}
}
\end{table}

\begin{figure}[htbp]
    \centering
        \includegraphics[width=0.7\linewidth]{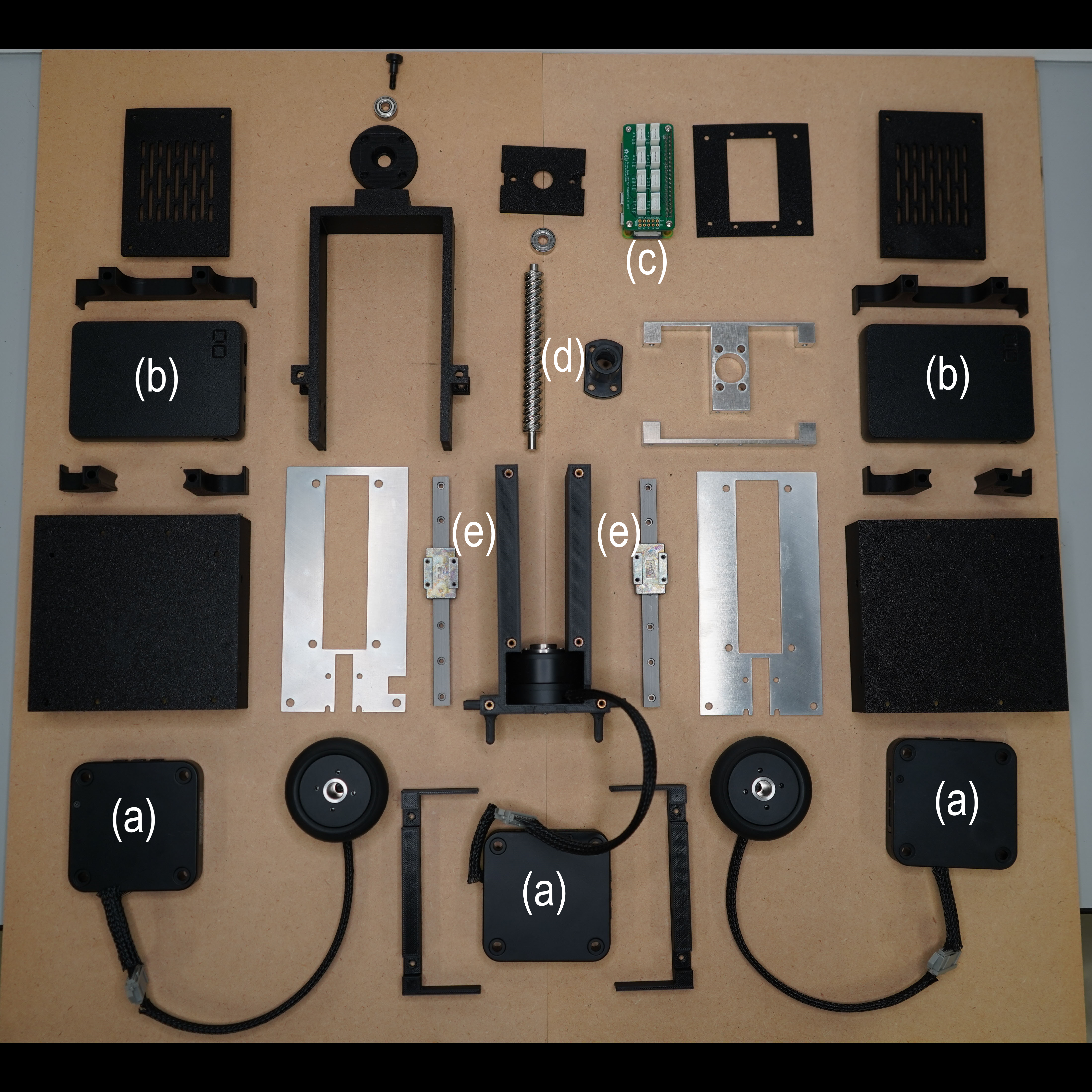}
    \caption*{Internal components of MOFU:(a) KM-1S-M4021, Keigan Inc.; (b) SMARTCOBY Pro SLIM, CIO Co., Ltd.; (c) Raspberry Pi Zero 2 W, Raspberry Pi Ltd; (d) MTSPW1220, MISUMI Group Inc.; (e) TK-04, Igus K.K. }
    \label{fig:mofu_internal}
\end{figure}

% MOFUの主な仕様を表\ref{tab:spec}に示す．垂直方向可変範囲，膨張収縮速度，連続稼働時間，総質量などを測定・確認し，機械系・電子系・制御系を統合したプロトタイプとして実装した．また，車輪についても毛糸カバー（SKI YARN Tenshi no Fur, Motohiro & Co., Ltd.）を装着し，外装と統一感を持たせつつ，機械的な印象をさらに低減した．
Figure~\hyperref[fig:mofu_internal]{S1} shows the internal components of MOFU.

% \begin{table}[htbp]
%   \centering
%   \caption{主なMOFUの仕様}
%   \label{tab:performance}
%   \begin{tabular}{lll}
%     \toprule
%     項目 & 値 & 測定条件・方法 \\
%     \midrule
%     外形寸法（タイヤ・フレーム含む高さ） & 215 -- 280 mm & 実測，静止時垂直方向最高点 \\
%     車輪の接地点間距離 & $90$ mm & 左右タイヤ接地点間距離\\
%     車輪径 & $58$ & ゴムタイヤ直径\\
%     総バッテリー容量 & 20000 mAh & \\
%     総重量 & 2091 g & バッテリー，外装含む \\
%     \bottomrule
%   \end{tabular}
%   \label{tab:spec}
% \end{table}

\subsection*{Modeling and validation of the expansion--contraction mechanism}
As mentioned earlier, the expansion--contraction motion of MOFU is based on the Jitterbug structure, which naturally produces rotational motion during deformation. 
% The expansion-contraction motion of MOFU is modeled based on the Jitterbug structure, which inherently involves rotational motion during deformation. 
This apparent rotation is minimized by the differential two-wheel drive, which compensates for the induced rotation, enabling MOFU to appear to expand and contract independently.
% To suppress the apparent rotation, the differential two-wheel drive was used to compensate for the induced rotational motion. 
% As a result, MOFU can appear to perform expansion-contraction alone.

The model equation was constructed based on the Jitterbug model proposed by Verheyen et al. \cite{Verheyen1989-kb}, with reference to the work of Chin et al. \cite{Chin2023-yg}. 
% The model equation was constructed based on the Jitterbug model proposed by Verheyen et al. \cite{Verheyen1989-kb}, with reference to the work of Chin et al. \cite{Chin2023-yg}. 
The vertical displacement $Z$ of the Jitterbug structure is expressed as a function of the rotation angle $\Theta$ of the top face relative to the base as follows:
% The vertical displacement $Z$ of the Jitterbug structure is expressed as a function of the rotation angle $\Theta$ of the top face relative to the base as follows:

\begin{equation}
Z(\Theta)=2\sqrt{r_x^2+r_y^2+r_z^2}+C
\label{eq1}
\end{equation}

where $r_x$, $r_y$, and $r_z$ are defined as

\begin{equation}
r_x(\mu)=R_A\cos\mu
\end{equation}
\begin{equation}
r_y(\mu)=R_A\sin\mu
\end{equation}
\begin{equation}
r_z(\mu)=\frac{R_A\cos\theta_{dh}\cos\mu+\sqrt{R_B^2-R_A^2\sin^2\mu}}{\sin\theta_{dh}}
\end{equation}

Here, $R_A$, $R_B$, and $\theta_{dh}$ are geometric parameters of the Jitterbug structure. Furthermore, $\mu$ is defined as

\begin{equation}
\mu=\mu_0+\theta
\end{equation}

where $\mu_0$ is calculated from $R_A$ and $R_B$ as follows:

\begin{equation}
\mu_0=\arcsin\frac{R_B}{R_A}
\end{equation}

The angle $\Theta$ is then related to $\theta$ as

\begin{equation}
\Theta=2\theta
\end{equation}

\begin{figure}[htbp]
\centering
\includegraphics[width=0.4\linewidth]{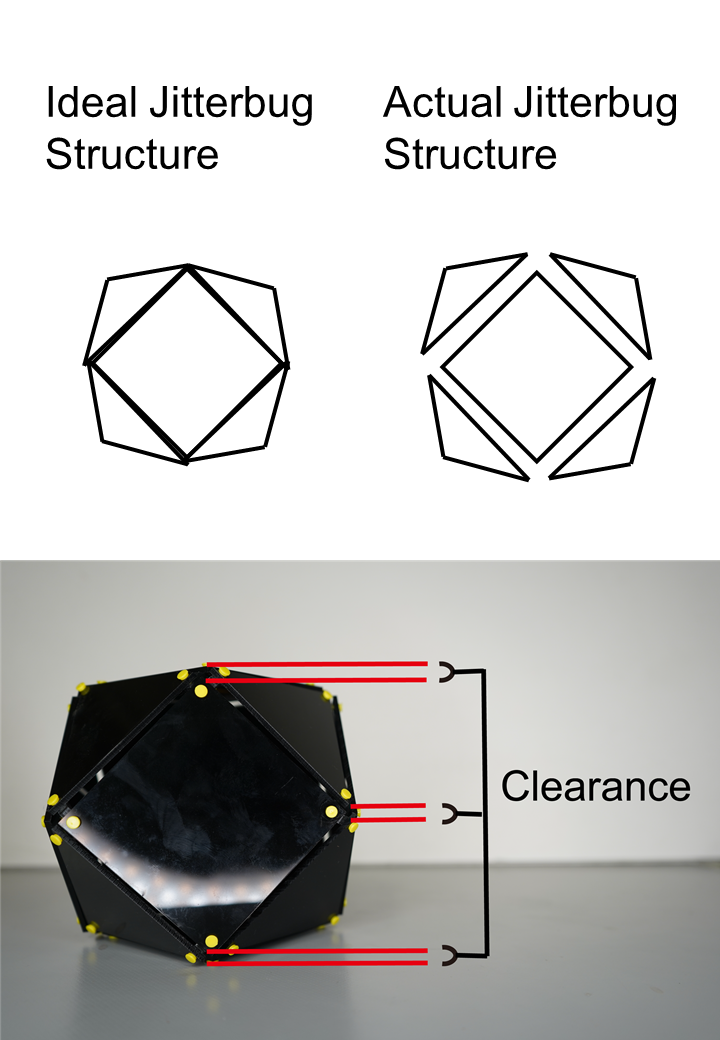}
\caption{Gap generated by the joints of the Jitterbug structure.}
\label{fig:clearance}
\end{figure}

Thus, once $\Theta$ is determined, $Z$ is uniquely obtained. 
% Thus, once $\Theta$ is determined, $Z$ is uniquely obtained.
In Eq.~\ref{eq1}, the term $C$ represents a clearance constant specific to MOFU's Jitterbug structure. 
% In Eq.~\ref{eq1}, the term $C$ represents a clearance constant specific to MOFU's Jitterbug structure. 
Although an ideal Jitterbug structure has no gaps between its faces, MOFU's joints introduce mechanical clearances (Fig.~\ref{fig:clearance}). 
% As shown in Fig.~\ref{fig:clearance}, while an ideal Jitterbug structure has no gaps between its faces, MOFU's joints introduce mechanical clearances.
This clearance is represented by the constant $C$. 
% This clearance was represented by the constant $C$.
The values of $R_A$, $R_B$, $\theta_{dh}$, and $C$ are summarized in Table~\ref{tab:parameters}.
% The values of $R_A$, $R_B$, $\theta_{dh}$, and $C$ are summarized in Table~\ref{tab:parameters}.

In practice, $Z$ is used as the control target for MOFU's expansion--contraction motion. 
% In practice, $Z$ is used as the control target for MOFU's expansion-contraction motion.
Therefore, the inverse function of Eq.~\ref{eq1} has to be computed. 
% Therefore, it is necessary to compute the inverse function of Eq.~\ref{eq1}.
However, a direct inverse calculation is not feasible because Eq.~\ref{eq1} is highly complex. 
% However, since Eq.~\ref{eq1} is highly complex, a direct inverse calculation is not feasible. 
Instead, the Jitterbug angle $\Theta$ was sampled at 45 evenly spaced points in the range $0$--$1.0$ rad, and the corresponding $Z$ values were precomputed. 
% Instead, the Jitterbug angle $\Theta$ was sampled at 45 evenly spaced points in the range of $0$ to $1.0$ rad, and the corresponding $Z$ values were precomputed. 
During control, when a target $Z$ was given, the corresponding $\Theta$ was selected from this table.
% During control, when a target $Z$ was given, the corresponding $\Theta$ was selected from this table.

\begin{table}[htbp]
\centering
\caption{Geometric parameters for the Jitterbug structure in MOFU.}
\label{tab:parameters}
\begin{tabular}{lll}
\hline
Parameter & Description & Value \\
\hline
$R_A$ & Radius associated with geometry A & $80\sqrt{2}/2 = 56.6$ mm \\
$R_B$ & Radius associated with geometry B & $80\sqrt{3}/3 = 46.2$ mm \\
$\theta_{dh}$ & Dihedral angle of the Jitterbug & $54.74^{\circ}$ (0.956 rad) \\
$C$ & Clearance constant & 13 mm \\
\hline
\end{tabular}
\end{table}

\begin{figure}[t]
\centering
% 1st row --------------------------------
\begin{subfigure}[b]{0.48\linewidth}
\centering
\includegraphics[width=\linewidth]{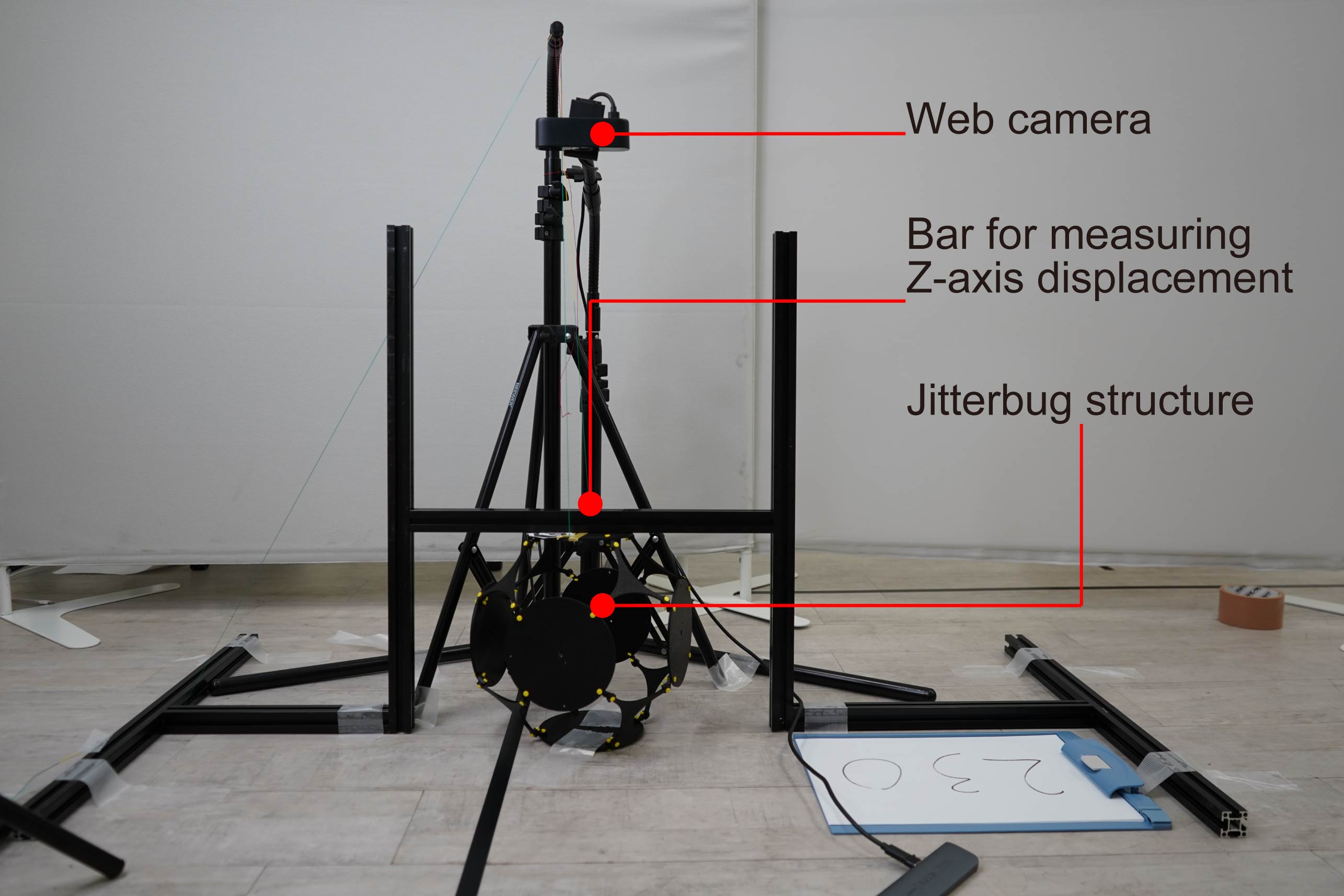}
\caption{}
\label{fig:measuringJBS}
\end{subfigure}\hfill
\begin{subfigure}[b]{0.48\linewidth}
\centering
\includegraphics[width=\linewidth]{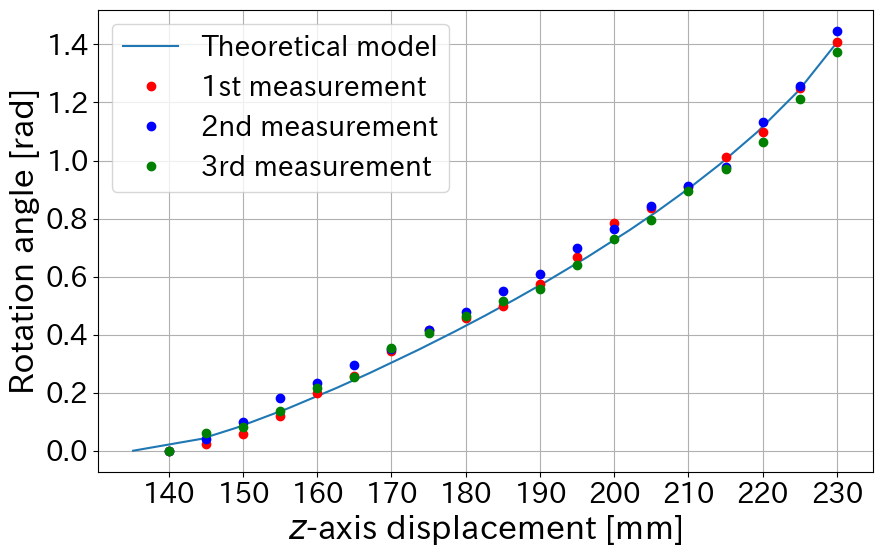}
\caption{}
\label{fig:graphJBS}
\end{subfigure}
\caption{(a) Experimental setup for measuring the Jitterbug structure. (b) Comparison between the model curve and measurement results.}
\label{fig:mofu2}
\end{figure}

To validate the model, we experimentally examined the relationship between the vertical displacement $Z$ and the rotation angle $\Theta$ of the Jitterbug structure. 
% To validate the model, the relationship between the vertical displacement $Z$ and the rotation angle $\Theta$ of the Jitterbug structure was experimentally examined. 
A webcam and ArUco markers \cite{Garrido-Jurado2014-vv} were used to measure the rotation angle, while the vertical displacement was measured three times using a jig. The experimental setup is shown in Fig.~\ref{fig:measuringJBS}.
% A webcam and ArUco markers \cite{Garrido-Jurado2014-vv} were used to measure the rotation angle, while the vertical displacement was measured three times using a jig. The experimental setup is shown in Fig.~\ref{fig:measuringJBS}.

The comparison between the measured results and the theoretical model is presented in Fig.~\ref{fig:graphJBS}. 
% The comparison between the measured results and the theoretical model is shown in Fig.~\ref{fig:graphJBS}.
The root mean squared errors (RMSE) of the estimated rotation angles for the three measurements were 0.025 rad, 0.038 rad, and 0.026 rad, respectively. 
% The root mean squared errors (RMSE) of the estimated rotation angles for the three measurements were 0.025 rad, 0.038 rad, and 0.026 rad, respectively.
In obtaining the model values, the corresponding $\Theta$ was selected using the aforementioned sampling method for the measured $Z$. 
% In obtaining the model values, the corresponding $\Theta$ was selected using the aforementioned sampling method for the measured $Z$. 
The small RMSE values confirmed the validity of the model.

\subsection*{Robot motions}
With the implementation of the design and control strategy explained in the preceding sections, MOFU was able to independently expand and contract and perform rotational motions using the differential two-wheel drive. 
% With the above implementation and control, MOFU was able to independently evaluate expansion-contraction and rotational motions by differential two-wheel drive. 
During an expansion--contraction motion with a vertical displacement of 65 mm, the robot base rotated by up to $28.6^{\circ}$, enabling evaluation of the two types of motion inherent to the Jitterbug structure---expansion--contraction and base rotation.
% During an expansion-contraction motion with a vertical displacement of 65 mm, the robot base rotated by up to $28.6^{\circ}$. Therefore, the two types of motion inherent to the Jitterbug structure---expansion-contraction and base rotation---could be evaluated.

\begin{table}[htbp]
\centering
\caption{Control parameters for the expansion--contraction motion.}
\begin{tabular}{ll}
\hline
Parameter & Value \\
\hline
Period & $6$ s \\
Motor rotation amplitude & $7 \pi$ rad (3.5 revolution)\\
Control frequency & $10$ Hz \\
\hline
\end{tabular}
\label{tab:motionparam}
\end{table}

The inverse function of the model is complex; hence, as mentioned earlier, the applicable range of the theoretical model (0--1.0 rad) was sampled at 45 points, and a correspondence table was created for motor control. 
% Since the inverse function of the model is complex, the applicable range of the theoretical model (0--1.0 rad) was sampled at 45 points, and a correspondence table was created for motor control.
This method enabled compensation for rotational motion induced by expansion--contraction, facilitating four motion conditions by combining the presence or absence of expansion--contraction and rotational motion (Table~\ref{tab:motionparam}). 
% This approach enabled the compensation of rotational motion induced by expansion-contraction, allowing four motion conditions to be realized by combining the presence or absence of expansion-contraction and rotational motions (Table~\ref{tab:motionparam}).
These conditions were used in the impression evaluation experiments described in the next section.
% These motion conditions were used in the impression evaluation experiments presented in the next section.

% 以上の実装と制御により，MOFUは膨張収縮動作と差動二輪による旋回動作を独立に評価できるようになった．垂直方向変位$65 \mathrm{mm}$の膨張収縮動作に伴い，ロボット底面は最大で$28.6^{\circ}$回転する．したがって，Jitterbug構造に内在する2種類の動き---すなわち膨張収縮と底面回転---について評価が可能である．

% \begin{table}[htbp]
%     \centering
%     \caption{膨張収縮動作における制御パラメータ}
%     \begin{tabular}{ll}
%     \hline
%     Parameter & Value \\
%     \hline
%     Period & $6$ (s)\\
%     Amplitude & $7 \pi$ (rad)\\
%     Control frequency & $10$ (Hz)\\
%      \hline
%     \end{tabular}
%     \label{tab:motionparam}
% \end{table}

% なお，モデルの逆関数が複雑であるため，理論モデルが適用可能な範囲（$0 \sim 1.0 \mathrm{rad}$）を45点サンプリングし，対応表を作成してモータ制御に用いた．この手法により，膨張収縮により生じる旋回を相殺でき，膨張収縮の有無と旋回の有無を組み合わせた4条件の動作が可能となった（表\ref{tab:motionparam}）．これらの条件を用いて，次章で示す印象評価実験を実施した．

% 導出した理論モデルについて，垂直変位から回転角度を求める逆関数は複雑であるため，理論モデルが適応可能な回転角度の範囲（$0 \sim 1.0 \mathrm{rad}$）を等間隔に45点サンプリングして$z$と$\mu$の対応表を作成し，モータ制御に用いた．
% この手法により膨張収縮動作に応じてロボットの差動2輪を制御し，膨張収縮機構により生じる旋回が相殺可能となった．
% したがって，MOFUのJitterbug構造本来の動きから膨張収縮の有無と旋回有無について4条件の動作を作成可能である．
% この動作条件を用いて，次章で示す印象評価実験を実施した．

\section*{Methods}
In this study, we conducted three experiments involving different participants to investigate the effect of whole-body expansion--contraction movements, which are volumetric motions, on the perception of animacy in robots.\
% In this study, we conducted three between-participants experiments to examine how whole-body expansion-contraction movements (volumetric motion) influence the perception of animacy in robots.

\begin{itemize}
\item \textbf{Experiment 1 (Single-robot condition):} Participants evaluated a single robot in a stationary state, under conditions with and without expansion--contraction and rotational motions.
% Participants evaluated a single robot presented in a stationary state, under conditions with and without expansion-contraction and rotational motions.
\item \textbf{Experiment 2 (Dual-robot condition):} Participants assessed their impressions when two robots were simultaneously displayed, both with and without expansion--contraction movements.
% Participants evaluated impressions when two robots were presented simultaneously, both with and without expansion-contraction movements.
\item \textbf{Experiment 3 (Locomotion condition):} Participants assessed the animacy of the robot when presented with locomotion alone or in conjunction with expansion--contraction.
% Participants evaluated the animacy of the robot when locomotion was presented alone or in combination with expansion-contraction.
\end{itemize}

These experiments were designed to test the following hypotheses:
\begin{enumerate}
\item Expansion--contraction increases perceived animacy compared with stationary or rotational motions alone.
% expansion-contraction increases perceived animacy compared with stationary or rotational motions alone.
\item Presenting two robots simultaneously increases perceived animacy compared with presenting a single robot.
% Presenting two robots simultaneously increases perceived animacy compared with a single robot.
\item Expansion--contraction further enhances animacy perception when combined with locomotion.
% expansion-contraction further enhances animacy perception when combined with locomotion.
\end{enumerate}

In addition to the a priori hypotheses, Experiment 2 included the same four motion conditions as Experiment 1 but with two robots, to examine generalization. 
% In addition to these a priori hypotheses, Experiment 2 also included the same four motion conditions as in Experiment 1 but with two robots, to examine generalization. 
Comparisons among these four dual-robot conditions were treated as exploratory analyses and are reported separately from the a priori predictions. 
The confirmatory test for the second prediction compares RM+EC with one robot versus RM+EC with two robots.
% Comparisons among these four dual-robot conditions were conducted as exploratory analyses and are reported separately from the a priori predictions; the confirmatory test related to the second prediction compares RM+EC with one robot versus RM+EC with two robots.

\subsection*{Participants}
We recruited participants online through a crowdsourcing platform (Yahoo! Crowdsourcing). 
% We recruited participants online through a crowdsourcing platform (Yahoo! Crowdsourcing). 
Recruitment was conducted in a single round for all three experiments (Experiment 1, Experiment 2, and Experiment 3), with participants randomly assigned to each experimental condition. 
% A single round of recruitment was conducted simultaneously for all three experiments (Experiment 1, Experiment 2, and Experiment 3), and participants were randomly assigned by the system to each experimental condition. 
Multiple participation across experiments was prohibited by the platform. In total, 561 participants took part, of whom 64 were excluded due to incomplete or invalid responses, leaving 498 valid participants (353 male, 140 female, 5 preferred not to answer; $M = 51.69$, $SD = 11.67$ years). 
% Multiple participation across experiments was prohibited by the platform. In total, 561 participants took part, of whom 64 were excluded due to incomplete or invalid responses, leaving 498 valid participants (353 male, 140 female, 5 preferred not to answer; $M = 51.69$, $SD = 11.67$ years). 
We provide information about the number of participants allocated to each experiment and the sample sizes that were calculated in advance in the following subsections. 
% The number of participants allocated to each experiment, along with the required sample sizes calculated in advance, are reported in the respective Participants subsections. 
For all statistical analyses conducted in this study, the significance level was set at $\alpha = 0.050$.
% For all statistical analyses conducted in this study, the significance level was set at $\alpha = 0.050$.

%全体の実験参加者数の報告
%クラウドソーシングサービス（Yahoo! Crowdsourcing）を通じてオンラインで実験参加者を募集した。実験1、実験2、実験3のすべてを対象として同時に1回の募集を行い、応募者はシステムによってランダムに各実験条件に割り当てられた。重複参加はプラットフォームの仕様により禁止されていた。合計561名が参加し、アンケート未回答や回答不備のあった64名を除外した結果、最終的に498名（男性353名、女性140名、回答を望まない5名；平均年齢 = $51.69 \pm 11.67$歳）のデータを解析に使用した。各実験に割り当てられた参加者数および事前に算出した必要サンプルサイズについては、それぞれの Participants サブセクションで記載する。155 male, 55 female; $M = 51.26$, $SD = 12.04$ years

\subsubsection*{Experiment 1 (Single-robot condition)}
The required sample size for this experiment was calculated using G*Power 3.1 software \cite{Faul2007-gd}. 
% The required sample size for this experiment was calculated using G*Power 3.1 software \cite{Faul2007-gd}. 
A $2 \times 2$ analysis of variance (ANOVA) was assumed with a medium effect size ($f = 0.25$), significance level $\alpha = 0.050$, and statistical power $1-\beta = 0.80$. 
% A $2 \times 2$ analysis of variance (ANOVA) was assumed with a medium effect size ($f = 0.25$), significance level $\alpha = 0.050$, and statistical power $1-\beta = 0.80$. 
Adequate power for detecting the main effects and the interaction was ensured by specifying the following parameters: numerator $df = 1$, denominator $df = N - 4$ (accounting for four groups), and number of groups $= 2 \times 2 = 4$. 
% To ensure adequate power for detecting the main effects and the interaction, the following parameters were specified: numerator $df = 1$, denominator $df = N - 4$ (accounting for four groups), and number of groups $= 2 \times 2 = 4$. 
The analysis indicated that a minimum of $N = 128$ participants was required. 
% The analysis indicated that a minimum of $N = 128$ participants was required. 
However, a larger number of participants was recruited in anticipation of invalid responses, resulting in a final valid sample of $N = 210$ participants. 
%  However, a larger number of participants was recruited in anticipation of invalid responses, resulting in a final valid sample of $N = 210$ participants. 
Of these, $n = 155$ were male, $n = 54$ were female and $n = 1$ was a participant who chose not to report their gender, with a mean age of $M = 51.26$, $SD = 12.04$ years. 
% Of these, $n = 155$ were male, $n = 54$ were female and $n = 1$ was a participant who chose not to report their gender, with a mean age of $M = 51.26$, $SD = 12.04$ years. 
These data were used for analysis.
% These data were used for analysis.

%本実験のサンプルサイズは，G*Power 3.1 ソフトウェア \cite{Faul2007-gd} を用いて算出した。$2 \times 2$ の分散分析 (ANOVA) を想定し，中程度の効果量 ($f = 0.25$)，有意水準 $\alpha = 0.050$，検出力 $1-\beta = 0.80$ を設定した。各主効果および交互作用の検定力を確保するため，以下のパラメータを用いた：Numerator $df = 1$，Denominator $df = N - 4$（4 群を考慮），Number of groups $= 2 \times 2 = 4$。この結果，必要なサンプルサイズは $N = 128$ であると算出された。ただし，無効回答の発生を見込んで多めに参加者を募集したため，最終的な有効回答は $N = 210$ 名であった。 その内訳は男性 $n = 155$，女性 $n = 55$，平均年齢 $M = 51.26$ 歳，標準偏差 $SD = 12.04$ 歳であり，これらのデータを分析に使用した。

\subsubsection*{Experiment 2 (Dual-robot condition)}
The required sample size for this experiment was calculated using G*Power 3.1 software \cite{Faul2007-gd}.
% The required sample size for this experiment was calculated using G*Power 3.1 software \cite{Faul2007-gd}.
A $2 \times 2$ analysis of variance (ANOVA) was assumed with a medium effect size ($f = 0.25$), significance level $\alpha = 0.050$, and statistical power $1-\beta = 0.80$.
% A $2 \times 2$ analysis of variance (ANOVA) was assumed with a medium effect size ($f = 0.25$), significance level $\alpha = 0.050$, and statistical power $1-\beta = 0.80$.
Sufficient power for detecting the main effects and their interaction was ensured by specifying the following parameters: numerator $df = 1$, denominator $df = N - 4$ (accounting for four groups), and number of groups $= 2 \times 2 = 4$.
% To ensure adequate power for detecting the main effects and the interaction, the following parameters were specified: numerator $df = 1$, denominator $df = N - 4$ (accounting for four groups), and number of groups $= 2 \times 2 = 4$.
The analysis indicated that a minimum of $N = 128$ participants was required.
% The analysis indicated that a minimum of $N = 128$ participants was required.
However, a larger sample was recruited to anticipate invalid responses, resulting in a final valid sample of $N = 207$ participants.
% However, a larger sample was recruited to anticipate invalid responses, resulting in a final valid sample of $N = 207$ participants. 
Of these, $n = 147$ were male and $n = 58$ were female and $n = 2$ were participants who chose not to report their gender, with a mean age of $M = 51.19$, $SD = 12.16$ years.
% Of these, $n = 147$ were male and $n = 58$ were female and $n = 2$ were participants who chose not to report their gender, with a mean age of $M = 51.19$, $SD = 12.16$ years. 
These data were used for analysis.
% These data were used for analysis.

\subsubsection*{Experiment 3 (Locomotion condition)}
The sample size for this experiment was determined using G*Power 3.1 software \cite{Faul2007-gd}.
% The sample size for this experiment was determined using G*Power 3.1 software \cite{Faul2007-gd}.
A priori power analysis was conducted assuming an independent-samples $t$ test (two-tailed), with a medium effect size (Cohen's $d = 0.50$), significance level $\alpha = 0.050$, and statistical power $1-\beta = 0.80$. 
% A priori power analysis was conducted assuming an independent-samples $t$ test (two-tailed), with a medium effect size (Cohen's $d = 0.50$), significance level $\alpha = 0.050$, and statistical power $1-\beta = 0.80$.
The group size allocation ratio was set to 1 (equal sample sizes). 
% The group size allocation ratio was set to 1 (equal sample sizes).
Based on these parameters, the required sample size was calculated to be $N = 128$ (64 per group). 
% Based on these parameters, the required sample size was calculated to be $N = 128$ (64 per group).
The analysis yielded $df = 126$, a critical $t$ value of 1.98, a noncentrality parameter $\delta = 2.83$, and an actual power of 0.801, indicating that at least 128 participants were necessary to ensure sufficient statistical power. 
% The analysis yielded $df = 126$, a critical $t$ value of 1.98, a noncentrality parameter $\delta = 2.83$, and an actual power of 0.801, indicating that at least 128 participants were necessary to ensure sufficient statistical power.
However, owing to platform constraints that necessitated an equal distribution of participants across all experimental conditions, the final valid sample size fell short of the target. 
% However, due to platform constraints that required equal allocation of participants across all experimental conditions, the final valid sample size fell short of this target.
After excluding invalid responses, the dataset comprised $N = 81$ participants (male $n = 51$, female $n = 28$, undisclosed $n = 2$), with a mean age of $M = 51.57$ years and a standard deviation of $SD = 9.26$ years. 
% After excluding invalid responses, the dataset consisted of $N = 81$ participants (male $n = 51$, female $n = 28$, undisclosed $n = 2$), with a mean age of $M = 51.57$ years and a standard deviation of $SD = 9.26$ years. 
These data were used for analysis; however,  the reduced sample size has to be considered when interpreting the results.
% These data were used for analysis, though the reduced sample size should be considered when interpreting the results.

%本実験のサンプルサイズは，G*Power 3.1 ソフトウェア \cite{Faul2007-gd} を用いて算出した。独立サンプルの $t$ 検定（両側検定）を想定し，中程度の効果量（Cohen's $d = 0.50$），有意水準 $\alpha = 0.050$，検出力 $1-\beta = 0.80$ を設定した。グループ間のサンプルサイズ比は 1（等しいサンプルサイズ）とした。その結果，各群に64名，合計 $N = 128$ 名が必要であると算出された。このとき，自由度 $df = 126$，臨界 $t$ 値 = 1.98，非中心パラメータ $\delta = 2.83$，実際の検出力は 0.801 であった。したがって，十分な検出力を確保するためには少なくとも128名の参加者が必要であることが示された。しかし，実際の有効回答は無効回答の除外により $N = 81$ 名となり，必要サンプルサイズを下回った。その内訳は男性 $n = XX$，女性 $n = XX$，平均年齢 $M = 51.57$ 歳，標準偏差 $SD = 9.26$ 歳であり，これらのデータを分析に使用した。

\subsection*{Materials}
%MOFUについて
The experiments employed videos of MOFU's movements, introduced in the preceding section, as stimuli. 
% The experiments used videos of MOFU's movements, introduced in the previous section, as stimuli. 
The experimental conditions were meticulously crafted by seamlessly integrating the motions generated by the Jitterbug structure with the differential two-wheel drive.
%  The experimental conditions were created by combining the motions generated by the Jitterbug structure and the differential two-wheel drive.
% 実験では，前のSectionで紹介したMOFUの動作を撮影した動画を刺激として用いた。Jitterbug構造と差動二輪によって実現される動作を組み合わせて条件を作成した．

Each video, lasting 20 s, commenced at 0 s with the robot in a contracted state. 
% All videos lasted 20 seconds.
% Each video began at 0 seconds with the robot in a contracted state.
At 15 s, a keyword appeared in the lower-right corner of the screen to confirm participant engagement. 
The video concluded at 20 s. 
% At 15 seconds, a keyword appeared in the lower-right corner of the screen to confirm that participants were watching the video, and the video ended at 20 seconds.
Below are the details of each motion.
Corresponding videos of each movement are provided as supporting materials and are available at Figshare (\href{https://figshare.com/s/ccdfe4d3c5a24cc05bc2}{S4 Video}).
Each video file is named according to the experimental condition it represents.
Conditions labeled “(2 robots)” in the manuscript are indicated as “×2” in the video filenames.
% The details of each motion are described below.
%Robot movement

The robot motion conditions used in the three experiments are summarized in Table~\ref{tab:experiment_conditions}. 
% The robot motion conditions used in the three experiments are summarized in Table~\ref{tab:experiment_conditions}.
In Experiment~1, a single robot was presented, and four conditions were compared by combining the presence or absence of rotation and expansion--contraction. 
% In Experiment~1, a single robot was presented, and four conditions were compared by combining the presence or absence of rotation and expansion-contraction.
The conditions and their abbreviations are as follows:
% The conditions and their abbreviations are as follows:

\begin{itemize}
\item \textbf{NBM (No Body Motion)}: Neither rotation nor expansion--contraction.
\item \textbf{RM (Rotational Motion)}: Rotation only (implemented by the differential two-wheel drive).
\item \textbf{EC (Expansion--Contraction)}: Expansion--contraction only (achieved by canceling the rotational component generated by the Jitterbug structure using the differential two-wheel drive).
\item \textbf{RM+EC (Rotational Motion + Expansion--Contraction)}: Rotation combined with expansion--contraction, realized solely by the Jitterbug structure.
\end{itemize}

In Experiment~2, the conditions were identical to those in Experiment~1, except that two robots were presented instead of one (denoted as ``(2 robots)'' after each condition name).
% In Experiment~2, the conditions were identical to those in Experiment~1, except that two robots were presented instead of one (denoted as ``(2 robots)'' after each condition name).

In Experiment~3, locomotion was introduced, and two conditions were compared: LOC (locomotion without expansion--contraction) and LOC+RM+EC (locomotion with rotation and expansion--contraction).
% In Experiment~3, locomotion was added, and two conditions were compared: LOC (locomotion without expansion-contraction) and LOC+RM+EC (locomotion with rotation and expansion-contraction).

% 3つの実験のために用意したロボットの動作条件を表\ref{tab:experiment_conditions}に示す。実験1では，ロボットを1台のみ提示し，Rotationの有無と膨張収縮（expansion-contraction, Exp.--Cont.）の有無を組み合わせた4条件を比較した。条件名と略称は以下の通りである。

% \begin{itemize}
% \item \textbf{NBM (No Body Motion)}: Rotation，Exp.--Cont.ともになし
% \item \textbf{RM (Rotational Motion)}: Rotationのみ（差動二輪によって実現）
% \item \textbf{EC (expansion-contraction)}: 膨張収縮のみ（Jitterbug構造の動作から生じる旋回成分を差動二輪で打ち消すことで実現）
% \item \textbf{RM+EC (Rotational Motion + expansion-contraction)}: Jitterbug構造のみで実現されるRotation+膨張収縮
% \end{itemize}

% 実験2では，提示するロボットを2台にした点を除き，条件は実験1と同じである（各条件名の後に ``(2 robots)'' と記載）。

% 実験3では，Locomotionを加え，RM+ECを伴わない場合と伴う場合の2条件を比較した（LOC および LOC+RM+EC）。

\begin{table}[htbp]
    \centering
    \caption{List of MOFU motion conditions. Abbreviations: NBM = No Body Motion, RM = Rotational Motion, EC = Expansion--Contraction (Exp.--Cont.), LOC = Locomotion. Note that comparisons among the four dual-robot motion conditions were exploratory and are reported separately from the confirmatory test of the two- versus one-robot comparison under EC.}
    \label{tab:experiment_conditions}
    \begin{tabular}{lcccc}
        \toprule
        Conditions & Robots & Rotation & Exp.--Cont. & Locomotion \\
        \midrule
        NBM & 1 & - & - & -\\
        RM & 1 & \ding{51} & - & -\\
        EC & 1 & - & \ding{51} & -\\
        RM+EC & 1 & \ding{51} & \ding{51} & -\\
        NBM (2 robots) & 2 & - & - & -\\
        RM (2 robots) & 2 & \ding{51} & - & -\\
        EC (2 robots) & 2 & - & \ding{51} & -\\
        RM+EC (2 robots) & 2 & \ding{51} & \ding{51} & -\\
        LOC & 1 & - & - & \ding{51}\\
        LOC+RM+EC & 1 & \ding{51} & \ding{51} & \ding{51}\\
        \bottomrule
    \end{tabular}
    \label{conditionstable}
\end{table}

%Questionnaire
Animacy was assessed using the Animacy subscale (Series II) of the Godspeed Questionnaire Series \cite{Bartneck2009-el}. 
% Animacy was assessed using the Animacy subscale (Series II) of the Godspeed Questionnaire Series \cite{Bartneck2009-el}. 
This scale is available in multiple languages \cite{Bartneck2023-op}. 
However, in this study, participants were restricted to Japanese speakers and responded to the Japanese version of the questionnaire. 
% This scale is available in multiple languages \cite{Bartneck2023-op}, and in this study, participants were restricted to Japanese speakers and responded to the Japanese version of the questionnaire.
Responses were collected on a 5-point Likert-type scale.
% Responses were collected on a 5-point Likert-type scale.

Participants were also required to answer a multiple-choice question (with 10 options) about a keyword presented in the middle of the video to prevent satisficing in the online survey. 
% To address potential satisficing in the online survey, participants were also required to answer a multiple-choice question (10 options) about a keyword presented in the middle of the video. 
Additionally, the Directed Questions Scale was appended to the end of the Godspeed Questionnaire Series. 
% In addition, the Directed Questions Scale was appended to the end of the Godspeed Questionnaire Series.
Furthermore, participants had the option to provide free-text comments about the robot's movements.
% Finally, participants were given the option to provide free-text comments about the robot's movements.

% アニマシーの評価には，Godspeed Questionnaire Series の下位尺度であるAnimacy尺度（Series II）を用いた \cite{Bartneck2009-el}。本尺度は多言語に対応しており \cite{Bartneck2023-op}，本研究では参加者を日本語話者に限定し，日本語版の質問項目を用いた。回答は5件法で収集した。

% さらに，オンラインアンケート特有のSatisficingへの対策として，動画の途中で提示されるキーワードを10択から回答させたほか，Godspeed Questionnaire Series の最後にDirected Questions Scaleを追加した。加えて，希望者のみ自由記述形式でロボットの動きに関するコメントを行うことができた。

% General information

% 使用ロボットは MOFU (Morphing Fluffy Unit) であることを明記。

% 直前の「Design and Development of MOFU」セクションを参照する形で簡単に紹介（詳細は再説明不要）。

% 刺激として用いたのは ロボットの動作を撮影した動画。

% 動画はオンライン実験で一貫して用いられたことを強調。

% 実験条件の一覧表を提示。

% 各Experiment (1, 2, 3) における条件を網羅的に示す（例：Stationary with/without expansion-contraction, Dual-robot with/without expansion-contraction, Locomotion alone/with expansion-contraction など）。

% General部分では条件名と組み合わせを表で示し、詳細な生成方法やパラメータは各Experimentのsubsubsectionに分けて説明。

% Questionnaire (Animacy evaluation)

% 評価尺度は Godspeed Questionnaire Series (Bartneck et al.) のAnimacy subscale。

% 項目数（5項目）と評価形式（7-point Likert scale）を明記。

% 得点化の方法（例：平均得点を算出し分析に用いた）を簡単に書く。

% Generalに記載し、Procedureには「参加者はこの質問紙に回答した」とだけ書けば十分。

\subsubsection*{Experiment 1 (Single-robot condition)}
In Experiment 1, participants were presented with four conditions: NBM, RM, EC, and RM+EC. 
% In Experiment 1, participants were presented with four conditions: NBM, RM, EC, and RM+EC. 
All robot movements followed a triangular waveform with a 6 s period. 
% The robot's movements in all conditions were controlled using a triangular waveform of 6 s period.
NBM represented the stationary state. 
% NBM represented the stationary state. 
RM reproduced only the rotational component generated by the Jitterbug structure using the differential two-wheel drive. 
% RM reproduced only the rotational component generated by the Jitterbug structure using the differential two-wheel drive.
EC presented expansion--contraction alone, with the rotational component inherent in the Jitterbug structure canceled out by the differential drive. 
% EC presented expansion-contraction alone, with the rotational component inherent in the Jitterbug structure canceled out by the differential drive.
RM+EC represented the condition in which both expansion--contraction and rotation occurred simultaneously through the Jitterbug structure.
% RM+EC represented the condition in which both expansion-contraction and rotation occurred simultaneously through the Jitterbug structure.

%実験1では、参加者にNBM, RM, EC, RM+ECの4条件の動画を提示した。すべての条件における動作は、6秒周期の三角波に従って制御された。NBMは静止状態を示し、RMはJitterbug構造によって生じる旋回成分のみを差動二輪で再現した。ECは膨張収縮動作のみを提示するため、Jitterbug構造に伴う旋回成分を差動二輪で打ち消した。RM+ECはJitterbug構造により膨張収縮と旋回が同時に表現される条件である。

\subsubsection*{Experiment 2 (Dual-robot condition)}
In Experiment 2, two robots were presented under the same four conditions as in Experiment 1: NBM (2 robots), RM (2 robots), EC (2 robots), and RM+EC (2 robots). 
% In Experiment 2, two robots were presented, with the same four conditions as in Experiment 1: NBM (2 robots), RM (2 robots), EC (2 robots), and RM+EC (2 robots). 
The expansion--contraction period for each robot was randomly varied between 5 and 7 s. 
% However, the period of expansion-contraction was randomly varied between 5 and 7 s for each robot. 
Reproducibility across videos was ensured by fixing the random seed so that the timing of expansion--contraction differed between the two robots in a reproducible manner. 
% To ensure reproducibility across videos, the random seed was fixed, so that the timing of expansion-contraction differed between the two robots in a reproducible manner.
This manipulation prevented full synchronization of their movements. 
% This manipulation was introduced to prevent their movements from becoming fully synchronized.
Comparisons among the four dual-robot conditions were pre-specified for descriptive characterization and treated as exploratory analyses. 
% Comparisons among these four dual-robot conditions were pre-specified for descriptive characterization and were treated as exploratory analyses.
Additionally, Experiment 2 included a confirmatory test of the prediction that presenting two robots would increase perceived animacy compared with a single robot under the EC condition.
% In addition, Experiment 2 provided the confirmatory test of our prediction that presenting two robots would increase animacy relative to one robot under EC.

% 実験2では、提示するロボットを2台にし、条件は実験1と同様にNBM（2 robots）, RM（2 robots）, EC（2 robots）, RM+EC（2 robots）を用いた。ただし、膨張収縮動作の周期は5～7秒の範囲からランダムに設定し、乱数シードを固定してすべての動画に同一系列を適用した。これは、2台のロボットの膨張収縮が完全に同期することを避けるための操作である。

\subsubsection*{Experiment 3 (Locomotion condition)}
In Experiment 3, participants were presented with two conditions: LOC and LOC+RM+EC. 
% In Experiment 3, participants were presented with two conditions: LOC and LOC+RM+EC. 
In both conditions, MOFU alternated between locomotion and stopping with a 2 s cycle, comprising 1.5 s of locomotion followed by 0.5 s of stopping. 
% In both conditions, MOFU alternated between locomotion and stopping with a cycle period of 2 s, consisting of 1.5 s of locomotion followed by 0.5 s of stopping.
In the LOC+RM+EC condition, expansion--contraction occurred only during locomotion phases and was withheld during stops. 
% In the LOC+RM+EC condition, expansion-contraction was executed only during locomotion phases and withheld during stops.
The expansion and contraction phases alternated across locomotion cycles, beginning with expansion on the first cycle.
% Expansion and contraction alternated across locomotion cycles, beginning with expansion on the first cycle.

% 実験3では、参加者にLOCとLOC+RM+ECの2条件を提示した。いずれの条件でも、MOFUは2秒周期で走行と停止を繰り返し、走行は1.5秒間、停止は0.5秒間とした。LOC+RM+EC条件では、走行中のみ膨張収縮動作を実行し、停止中は膨張収縮を行わなかった。膨張収縮は走行のたびに膨張と収縮を交互に切り替えるよう設定した。

\subsection*{Procedure}
Participants provided informed consent via an online form and then watched a single 20-s video. 
% Participants provided informed consent through an online form and then viewed a single 20-second video.
After viewing the video, they answered a multiple-choice question (10 options) about a keyword appearing in the latter part of the video. 
% After watching the video, they answered a multiple-choice question (10 options) about a keyword that appeared in the latter part of the video.
Next, they completed the six items of the Animacy subscale of the Godspeed Questionnaire Series, followed by one item from the Directed Questions Scale. 
% They then completed the six items of the Animacy subscale of the Godspeed Questionnaire Series, followed by one item from the Directed Questions Scale.
Participants were then given the option to provide free-text comments about the robot's movements. 
% Subsequently, participants had the option to provide free-text comments about the robot's movements. 
Those who correctly answered the keyword question received a monetary reward of 20 yen. 
% Participants who correctly answered the keyword question received a monetary reward of 20 yen.
Each participant viewed only one video condition and could not participate in multiple conditions.
% Each participant responded to only one video condition and was not allowed to participate in multiple conditions.

% 参加者はオンラインフォームでインフォームド・コンセントに同意した後、20秒間の動画を1本視聴した。動画視聴後、動画後半に提示されたキーワードを10択から回答し、その後 Godspeed Questionnaire Series のアニマシー尺度6項目と Directed Questions Scale 1項目に回答した。続いて、希望者のみ自由記述でロボットの動きに関するコメントを行った。正答した参加者には謝金が支払われた。各参加者は1つの条件動画のみに回答し、複数の条件に参加することはできなかった。

\subsection*{Ethical considerations}
The study protocol was approved by the Ethics Committee of the University of Electro-Communications (No. H23064(4)) and by the Sony Bioethics Committee (No. 23-R-0040). 
% The study protocol was approved by the Ethics Committee of the University of Electro-Communications (No. H23064(4)) and by the Sony Bioethics Committee (No. 23-R-0040). 
All participants provided written informed consent before taking part in the experiment. The online questionnaire was released on December 3, 2024, and responses were collected and finalized on the same day.
% All participants provided written informed consent before taking part in the experiment.

\section*{Results}
In this study, we investigated participants’ perceptions of animacy in response to the movements of MOFU, a robot capable of whole-body expansion and contraction. 
% In this study, we evaluated participants' impressions of animacy in response to the movements of MOFU, a robot capable of whole-body expansion and contraction. 
The findings from three experiments are presented below.
% The results of the three experiments are presented below.

% 本研究では，開発した全身で膨張収縮が可能なロボット MOFU の動きについて，Animacy に関する印象評価を行った．以下に3つの実験の結果をそれぞれ示す．

\subsection*{Experiment 1 (Single-robot condition)}
\begin{figure}[htbp]
\centering
\includegraphics[width=0.5\textwidth]{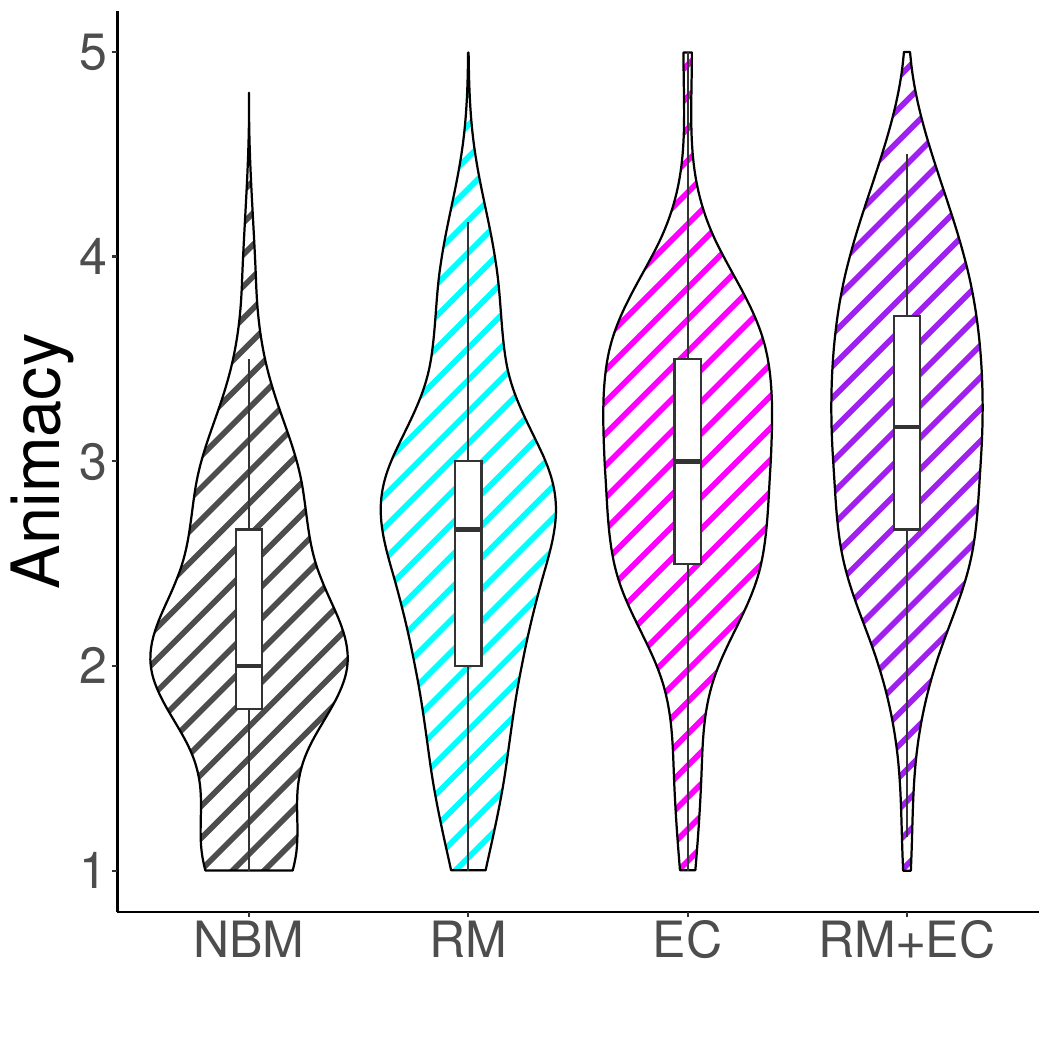}
\caption{Comparison of animacy across the single-robot conditions. The y-axis shows the animacy score, computed for each participant as the mean of the six items of the Godspeed Animacy subscale (Series II), rated on a 1--5 Likert-type scale (1 = low, 5 = high). The boxes represent the distribution across participants (median, quartiles, and range).}
\label{fig4}
\end{figure}

The median animacy score was lowest in the NBM condition, followed by RM, EC, and RM+EC (Figure~\ref{fig4}).
% As shown in Figure~\ref{fig4}, the median animacy score was lowest in the NBM condition, followed by RM, EC, and RM+EC.
% RCommander.txt 244-255
The descriptive statistics for each condition are presented in Table~\ref{tab:basic_exp1}.
% The descriptive statistics for each condition are presented in Table~\ref{tab:basic_exp1}.

\begin{table}[htbp]
  \centering
  \caption{Descriptive statistics of animacy scores by condition in Experiment 1}
  \label{tab:basic_exp1}
  \begin{tabular}{lccc}
    \toprule
    Condition & Participants & Mean & SD \\
    \midrule
    NBM & 52 & 2.12 & 0.72 \\
    RM & 51 & 2.96 & 0.74 \\
    EC & 55 & 2.61 & 0.79 \\
    RM+EC & 52 & 3.17 & 0.75 \\
    \bottomrule
  \end{tabular}
\end{table}

Prior to statistical testing, assumptions were checked. 
% Prior to statistical testing, assumptions were checked.
Shapiro--Wilk tests indicated normality in all conditions (NBM: $W(52) = 0.966, p = 0.137$; RM: $W(51) = 0.976, p = 0.375$; EC: $W(55) = 0.974, p = 0.289$; RM+EC: $W(52) = 0.979, p = 0.496$). 
% Shapiro--Wilk tests indicated normality in all conditions (NBM: $W(52) = 0.966, p = 0.137$; RM: $W(51) = 0.976, p = 0.375$; EC: $W(55) = 0.974, p = 0.289$; RM+EC: $W(52) = 0.979, p = 0.496$). 
Levene's test confirmed homogeneity of variance across conditions ($F(3, 206) = 0.218, p = 0.884$). Thus, the assumptions for ANOVA were met.
% Levene's test confirmed homogeneity of variance across conditions ($F(3, 206) = 0.218, p = 0.884$). Thus, the assumptions for ANOVA were met.
% RCommander.txt 155-200

A two-way ANOVA was conducted to investigate the effects of rotational motion, expansion--contraction, and their interaction. 
% A two-way ANOVA was conducted to examine the effects of the rotational motion and the expansion-contraction, as well as their interaction. 
Details of the ANOVA are shown in Table~\ref{tab:anova_type3_exp1}. 
% Details of the ANOVA are shown in Table~\ref{tab:anova_type3_exp1}. 
The main effect of the rotational motion was significant, $F(1, 206) = 11.236, p < 0.001$. 
The main effect of expansion--contraction was also significant, $F(1, 206) = 46.208, p < 0.001$. 
% The main effect of the rotational motion was significant, $F(1, 206) = 11.236, p < 0.001$. The main effect of expansion-contraction was also significant, $F(1, 206) = 46.208, p < 0.001$. 
In contrast, the interaction between the rotational motion and expansion--contraction was not significant, $F(1, 206) = 1.836, p = 0.177$.
% In contrast, the interaction between the rotational motion and the expansion-contraction was not significant, $F(1, 206) = 1.836, p = 0.177$.

\sisetup{
  table-number-alignment = center,
  group-separator = {,},
}

\begin{table}[htbp]
  \centering
  \caption{Results of the two-way ANOVA in Experiment 1}
  \label{tab:anova_type3_exp1}
  \begin{tabularx}{\linewidth}{
    >{\raggedright\arraybackslash}X
    S[table-format=4.2] % Sum Sq.
    S[table-format=3.0] % df
    S[table-format=4.2] % MS
    S[table-format=4.3] % F
    S[table-format=1.3] % p
    S[table-format=1.3] % partial eta^2
  }
    \toprule
    \multicolumn{1}{l}{Source} &
    \multicolumn{1}{c}{Sum Sq.} &
    \multicolumn{1}{c}{df} &
    \multicolumn{1}{c}{Mean Square} &
    \multicolumn{1}{c}{F} &
    \multicolumn{1}{c}{Sig.} &
    \multicolumn{1}{c}{$\eta^2_{\text{p}}$} \\
    \midrule
    {Corrected Model} & 33.17  & 3   & 11.06   & 19.766  & \multicolumn{1}{c}{$<0.001$} & 0.224 \\
    {Intercept}       & 1547.62& 1   & 1547.62 & 2766.753& \multicolumn{1}{c}{$<0.001$} & \multicolumn{1}{c}{--} \\
    {RM}              & 6.29   & 1   & 6.29    & 11.236  & \multicolumn{1}{c}{0.001}    & 0.050 \\
    {EC}              & 25.85  & 1   & 25.85   & 46.208  & \multicolumn{1}{c}{$<0.001$} & 0.180 \\
    {RM$\times$EC}    & 1.03   & 1   & 1.03    & 1.836   & 0.177                          & 0.009 \\
    \midrule
    {Error}           & 115.23 & 206 & 0.559   & \multicolumn{1}{c}{} & \multicolumn{1}{c}{} & \multicolumn{1}{c}{} \\
    {Total}           & 1696.02& \multicolumn{1}{c}{} & \multicolumn{1}{c}{} & \multicolumn{1}{c}{} & \multicolumn{1}{c}{} & \multicolumn{1}{c}{} \\
    {Corrected Total} & 148.40 & 209 & \multicolumn{1}{c}{} & \multicolumn{1}{c}{} & \multicolumn{1}{c}{} & \multicolumn{1}{c}{} \\
    \bottomrule
  \end{tabularx}
\end{table}

% RCommander.txt 211-217

Given the lack of a significant interaction, post hoc comparisons were conducted using Tukey’s HSD test to examine pairwise differences among the four conditions. 
% Given the no significant interaction, post hoc comparisons were performed using Tukey’s HSD test to examine pairwise differences among the four conditions.
The results indicated that the rotational motion condition (RM) had significantly higher animacy ratings than the baseline (NBM) ($p_{\mathrm{adj}} = 0.006$). 
% The results showed that the rotational motion condition (RM) had significantly higher animacy ratings than the baseline (NBM) ($p_{\mathrm{adj}} = 0.006$). 
The expansion--contraction condition (EC) also had significantly higher animacy ratings than the baseline (NBM) ($p_{\mathrm{adj}} < 0.001$). 
% The expansion-contraction condition (EC) also had significantly higher animacy ratings than the baseline (NBM) ($p_{\mathrm{adj}} < 0.001$). 
Similarly, the combined condition with both rotation and expansion--contraction (RM+EC) was significantly higher than the baseline (NBM) ($p_{\mathrm{adj}} < 0.001$). 
% Likewise, the combined condition with both rotation and expansion-contraction (RM+EC) was significantly higher than the baseline (NBM) ($p_{\mathrm{adj}} < 0.001$). 
In addition, the combined condition (RM+EC) showed significantly higher animacy than the rotational motion condition (RM) ($p_{\mathrm{adj}} = 0.001$). 
% In addition, the combined condition (RM+EC) showed significantly higher animacy than the rotational motion condition (RM) ($p_{\mathrm{adj}} = 0.001$).
No significant differences was found between the rotational motion (RM) and expansion--contraction (EC) conditions ($p_{\mathrm{adj}} = 0.072$) or between the expansion--contraction-only (EC) and combined (RM+EC) conditions ($p_{\mathrm{adj}} = 0.485$).
% No significant differences was found between conditions the rotational motion (RM) and the expansion-contraction (EC) ($p_{\mathrm{adj}} = 0.072$) or between conditions the expansion-contraction-only (EC) and the combined (RM+EC) ($p_{\mathrm{adj}} = 0.485$).

% RCommander.txt 231-238

These results suggest that both rotational motion and expansion--contraction enhanced animacy perception. 
Moreover, while the combined condition increased perceived animacy compared to rotation alone, it was not significantly different from expansion--contraction.

\subsection*{Experiment 2 (Dual-robot condition)}
\begin{figure}[htbp]
\centering
\includegraphics[width=0.5\textwidth]{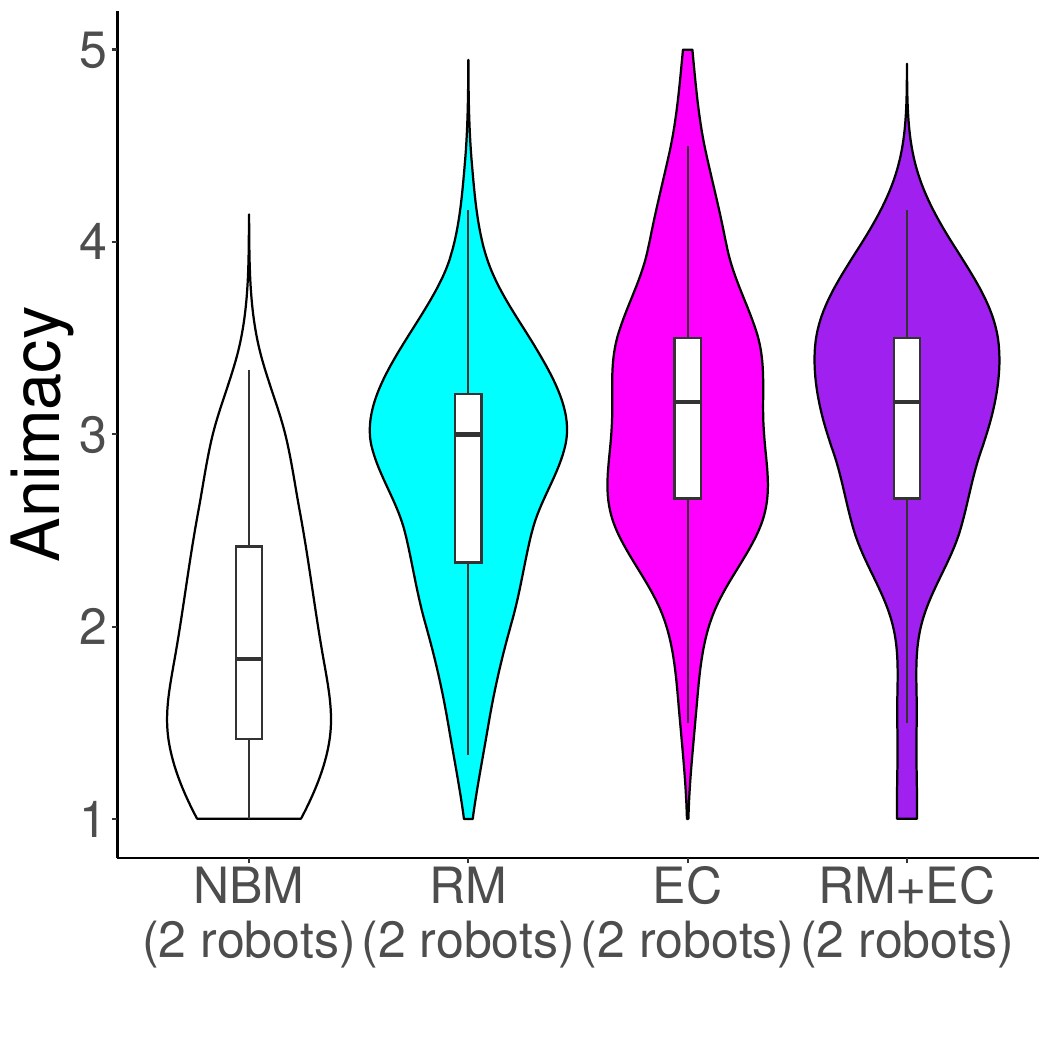}
\caption{(Exploratory) Comparison of animacy across the dual-robot conditions. The y-axis shows the animacy score, computed for each participant as the mean of the six items of the Godspeed Animacy subscale (Series II), rated on a 1--5 Likert-type scale (1 = low, 5 = high). The boxes represent the distribution across participants (median, quartiles, and range).}
\label{fig5}
\end{figure}

Exploratory analyses were conducted to compare animacy ratings among the four dual-robot conditions. 
% Exploratory analyses were conducted to compare animacy ratings among the four dual-robot conditions. 
The median animacy score was lowest in the NBM (2 robots) condition, followed by the RM (2 robots) condition, while the RM+EC (2 robots) and EC (2 robots) conditions showed identical median values (Figure~\ref{fig5}).
% As shown in Figure~\ref{fig5}, the median animacy score was lowest in the NBM (2 robots) condition, followed by RM (2 robots), RM+EC (2 robots), and EC (2 robots).
The descriptive statistics for each condition are listed in Table~\ref{tab:desc_kw_exp2}.
% The descriptive statistics for each condition are presented in Table~\ref{tab:desc_kw_exp2}.

\begin{table}[htbp]
  \centering
  \caption{Descriptive statistics of animacy scores by condition in Experiment 2}
  \label{tab:desc_kw_exp2}
  \begin{tabular}{lccc}
    \toprule
    Condition & Participants & Median & IQR \\
    \midrule
    NBM (2 robots) & 51 & 1.83 & 1.00 \\
    RM (2 robots)  & 52 & 3.00 & 0.88 \\
    EC (2 robots)  & 51& 3.17 & 0.83 \\
    RM+EC (2 robots)  & 53 & 3.17 & 0.83 \\
    \bottomrule
  \end{tabular}
\end{table}

Shapiro--Wilk tests were performed to ensure the validity of the statistical test. 
% To check the assumptions for statistical testing, Shapiro--Wilk tests were conducted. 
The assumption of normality was rejected for the NBM (2 robots) condition ($W(51) = 0.942, p = 0.015$) and the RM+EC (2 robots) condition ($W(53) = 0.918, p = 0.001$). 
% The assumption of normality was rejected for the NBM (2 robots) condition ($W(51) = 0.942, p = 0.015$) and the RM+EC (2 robots) condition ($W(53) = 0.918, p = 0.001$). 
In contrast, normality was confirmed for the RM (2 robots) condition ($W(52) = 0.968, p = 0.176$) and the EC (2 robots) condition ($W(51) = 0.986, p = 0.789$). 
% In contrast, normality was confirmed for the RM (2 robots) condition ($W(52) = 0.968, p = 0.176$) and the EC (2 robots) condition ($W(51) = 0.986, p = 0.789$).
After applying Holm’s correction, the violation of normality remained for NBM (2 robots) ($p_{\mathrm{adj}} = 0.045$) and RM+EC (2 robots) ($p_{\mathrm{adj}} = 0.006$), whereas normality was retained for RM (2 robots) ($p_{\mathrm{adj}} = 0.353$) and EC (2 robots) ($p_{\mathrm{adj}} = 0.789$).
% After applying Holm’s correction, the violation of normality remained for NBM (2 robots) ($p_{\mathrm{adj}} = 0.045$) and RM+EC (2 robots) ($p_{\mathrm{adj}} = 0.006$), whereas normality was retained for RM (2 robots) ($p_{\mathrm{adj}} = 0.353$) and EC (2 robots) ($p_{\mathrm{adj}} = 0.789$).
% RCommander.txt 304-343

As the assumption of normality was not met in certain conditions, a non-parametric Kruskal--Wallis test was conducted. 
% Because the assumption of normality was not satisfied in some conditions, a non-parametric Kruskal--Wallis test was performed.
The results revealed a significant difference among conditions, $\chi^2(3) = 66.844, p < 0.001$. 
% The results revealed a significant difference among conditions, $\chi^2(3) = 66.844, p < 0.001$. 
Post hoc pairwise comparisons using Dunn’s test with Bonferroni correction indicated that the NBM (2 robots) condition was rated significantly lower than RM (2 robots) ($Z = -5.047, p_{\mathrm{adj}} < 0.001$), EC (2 robots) ($Z = -7.030, p_{\mathrm{adj}} < 0.001$), and RM+EC (2 robots) ($Z = -7.139, p_{\mathrm{adj}} < 0.001$). 
% Post hoc pairwise comparisons using Dunn’s test with Bonferroni correction indicated that the NBM (2 robots) condition was rated significantly lower than RM (2 robots) ($Z = -5.047, p_{\mathrm{adj}} < 0.001$), EC (2 robots) ($Z = -7.030, p_{\mathrm{adj}} < 0.001$), and RM+EC (2 robots) ($Z = -7.139, p_{\mathrm{adj}} < 0.001$).
In contrast, no significant differences were observed between RM (2 robots) and EC (2 robots) ($Z = -2.017, p_{\mathrm{adj}} = 0.131$), between RM (2 robots) and RM+EC (2 robots) ($Z = -2.078, p_{\mathrm{adj}} = 0.113$), or between EC (2 robots) and RM+EC (2 robots) ($Z = -0.042, p_{\mathrm{adj}} = 1.000$).
% In contrast, no significant differences were observed between RM (2 robots) and EC (2 robots) ($Z = -2.017, p_{\mathrm{adj}} = 0.131$), between RM (2 robots) and RM+EC (2 robots) ($Z = -2.078, p_{\mathrm{adj}} = 0.113$), or between EC (2 robots) and RM+EC (2 robots) ($Z = -0.042, p_{\mathrm{adj}} = 1.000$).

% RCommander.txt 345-382

\begin{figure}[htbp]
\centering
\includegraphics[width=0.5\textwidth]{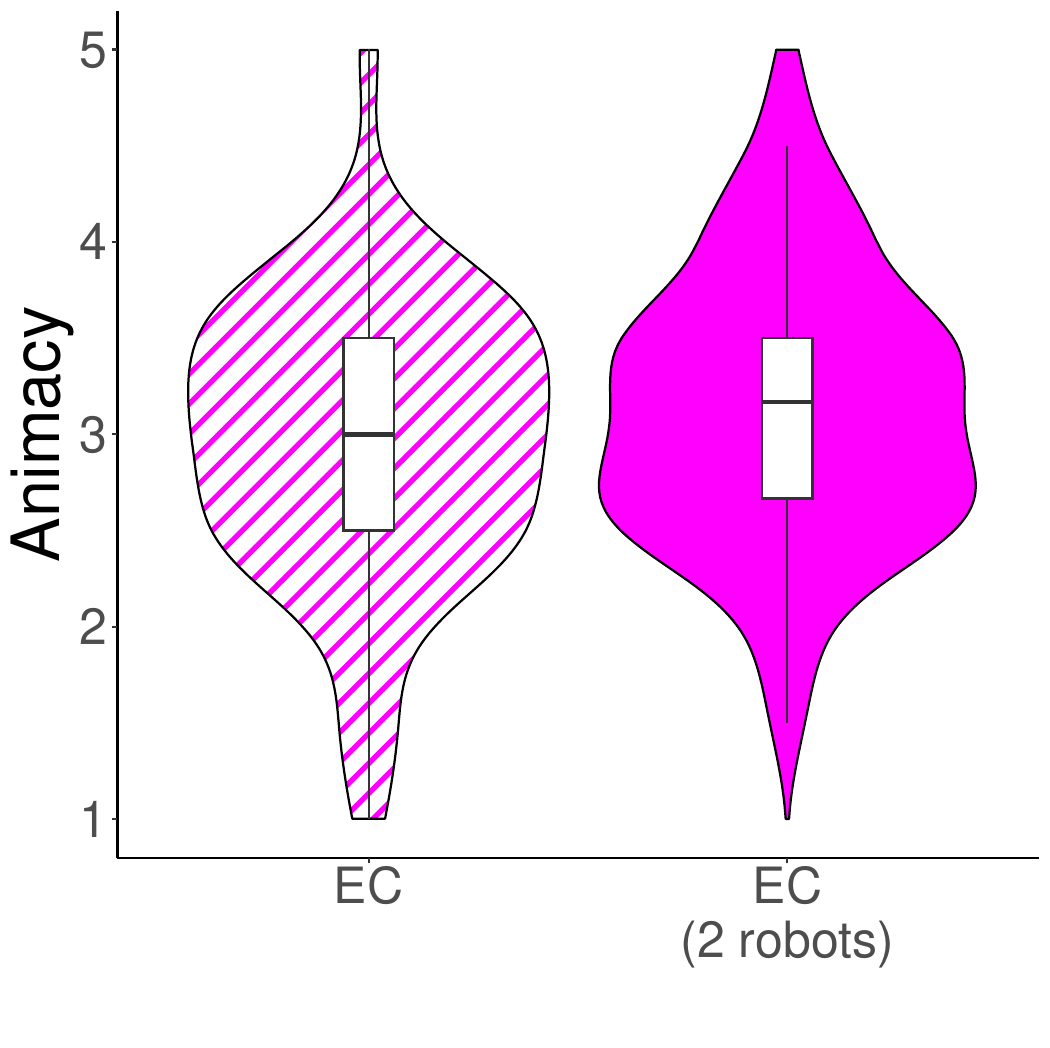}
\caption{Comparison of animacy between single-robot and dual-robot EC conditions. The y-axis shows the animacy score, computed for each participant as the mean of the six items of the Godspeed Animacy subscale (Series II), rated on a 1--5 Likert-type scale (1 = low, 5 = high). The boxes represent the distribution across participants (median, quartiles, and range).}
\label{fig7}
\end{figure}

To test our hypothesis, we compared the EC condition with EC (2 robots) (Figure~\ref{fig7}). 
% To test our hypothesis, we compared the EC condition with EC (2 robots) (Figure~\ref{fig7}).
Levene’s test indicated no significant difference in variances ($F(1, 104) = 0.008, p = 0.929$). 
% Levene’s test indicated no significant difference in variances ($F(1, 104) = 0.008, p = 0.929$). 
An independent-samples $t$-test assuming equal variances revealed no significant difference between the two conditions, $t(104) = -1.173, p = 0.244$. 
% An independent-samples $t$-test assuming equal variances revealed no significant difference between the two conditions, $t(104) = -1.173, p = 0.244$.
The 95\% confidence interval was [-0.441, 0.113], indicating that the mean difference between the two conditions was not statistically significant. 
% The 95\% confidence interval was [-0.441, 0.113], indicating that the mean difference between the two conditions was not statistically significant.
These results do not support the hypothesis that presenting two robots simultaneously increases perceived animacy compared with a single robot.

\subsection*{Experiment 3 (Locomotion condition)}
\begin{figure}[htbp]
\centering
\includegraphics[width=0.5\textwidth]{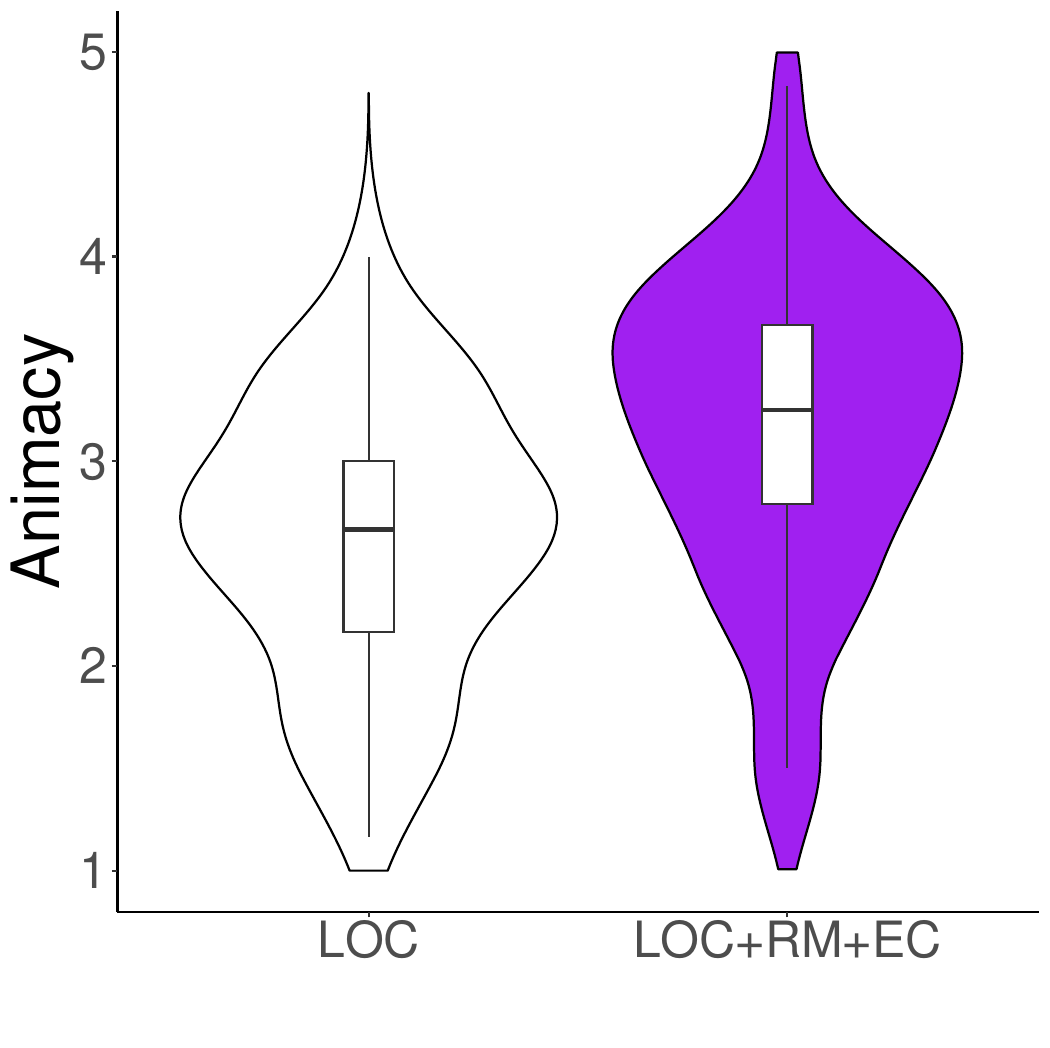}
\caption{Comparison of animacy across the locomotion conditions. The y-axis shows the animacy score, computed for each participant as the mean of the six items of the Godspeed Animacy subscale (Series II), rated on a 1--5 Likert-type scale (1 = low, 5 = high). The boxes represent the distribution across participants (median, quartiles, and range).}
\label{fig6}
\end{figure}

The median animacy score was higher in the condition with expansion--contraction during locomotion (LOC+RM+EC) than in the condition without expansion--contraction (LOC) (Figure~\ref{fig6}).
% As shown in Figure~\ref{fig6}, the median animacy score was higher in the condition with expansion-contraction during locomotion (LOC+RM+EC) than in the condition without expansion-contraction (LOC).
The descriptive statistics for each condition are presented in Table~\ref{tab:desc_ttest_exp3}.
% The descriptive statistics for each condition are presented in Table~\ref{tab:desc_ttest_exp3}.

\begin{table}[htbp]
  \centering
  \caption{Descriptive statistics of animacy scores by condition in Experiment 3}
  \label{tab:desc_ttest_exp3}
  \begin{tabular}{lccc}
    \toprule
    Condition & Participants & Mean & SD \\
    \midrule
    LOC+RM+EC & 40 & 3.16 & 0.77 \\
    LOC & 41 & 2.62 & 0.68 \\
    \bottomrule
  \end{tabular}
\end{table}

Shapiro--Wilk tests were conducted to check the assumptions for statistical testing. 
% To check the assumptions for statistical testing, Shapiro--Wilk tests were conducted.
Normality was confirmed for both conditions (LOC+RM+EC: $W(40) = 0.971, p = 0.389$; LOC: $W(41) = 0.977, p = 0.581$). 
% Normality was confirmed for both conditions (LOC+RM+EC: $W(40) = 0.971, p = 0.389$; LOC: $W(41) = 0.977, p = 0.581$).
Levene’s test also confirmed homogeneity of variance between the two conditions ($F(1, 79) = 0.412, p = 0.523$).
% Levene’s test also confirmed homogeneity of variance between the two conditions ($F(1, 79) = 0.412, p = 0.523$).

Accordingly, an independent-samples $t$-test was performed. 
% Accordingly, an independent-samples $t$-test was performed. 
The results showed that the mean animacy score in the LOC+RM+EC condition ($M = 3.158$) was significantly higher than in the LOC condition ($M = 2.618$), $t(79) = 3.359, p = 0.001$. 
% The results showed that the mean animacy score in the LOC+RM+EC condition ($M = 3.158$) was significantly higher than in the LOC condition ($M = 2.618$), $t(79) = 3.359, p = 0.001$. 
The 95\% confidence interval for the difference between the two means was [0.220, 0.861], indicating a statistically significant difference. 
% The 95\% confidence interval was [0.220, 0.861], indicating that the difference between the two means was statistically significant.
This suggests that incorporating expansion--contraction during locomotion significantly enhances the perceived animacy of MOFU.
% This result suggests that incorporating expansion-contraction during locomotion significantly enhances the perceived animacy of MOFU.

\begin{comment}
!!!!!!!!!!!!!!!!!!!!!!!!!!!!!!!!新バージョン!!!!!!!!!!!!!!!!!!!!!!!!!!!!!!
% 図\ref{fig6}に示すように，箱ひげ図の中央値を見ると，アニマシー評価は走行中に膨張収縮動作を行う条件（LOC+RM+EC）の方が，膨張収縮を行わない条件（LOC）よりも高い傾向が見られた。

% 統計分析の前提条件を確認するために Shapiro-Wilk 検定を行った。その結果，いずれの条件でも正規性が確認された（LOC+RM+EC: $W(40) = 0.971, p = 0.389$；LOC: $W(41) = 0.977, p = 0.581$）。また，Levene 検定により分散の等質性も確認された（$F(1, 79) = 0.412, p = 0.523$）。

% これらの結果から，独立二標本 $t$ 検定を実施した。その結果，LOC+RM+EC 条件の平均値（$M = 3.158$）は LOC 条件（$M = 2.618$）よりも有意に高いことが示された（$t(79) = 3.359, p = 0.001$）。95\%信頼区間は [0.220, 0.861] であり，両条件の平均値の差が統計的に有意であることを支持した。この結果は，走行中に膨張収縮動作を伴うことで MOFU のアニマシー評価が有意に高まることを示唆している。
!!!!!!!!!!!!!!!!!!!!!!!!!!!!!!!!!!!!!!!!!!!!!!!!!!!!!!!!!!!!!!!!!!!!!!!!!
\end{comment}

% 図\ref{fig6}に示すように，箱ひげ図の中央値を見ると，アニマシー評価は走行中に膨張収縮動作を行う条件（LOC+RM+EC）の方が，膨張収縮を行わない条件（LOC）よりも高い傾向が見られた。

% 統計分析の前提条件を確認するために Shapiro-Wilk 検定を行った。その結果，いずれの条件でも正規性が確認された（LOC+RM+EC: $W(n) = 0.97, p = 0.39$；LOC: $W(n) = 0.98, p = 0.58$）。また，Levene 検定により分散の等質性も確認された（$F(1, 79) = 0.41, p = 0.52$）。

% これらの結果から，独立二標本 $t$ 検定を実施した。その結果，LOC+RM+EC 条件の平均値（$M = 3.16$）は LOC 条件（$M = 2.62$）よりも有意に高いことが示された（$t(79) = 3.36, p < 0.001$）。95\%信頼区間は [0.22, 0.86] であり，両条件の平均値の差が統計的に有意であることを支持した。この結果は，走行中に膨張収縮動作を伴うことで MOFU のアニマシー評価が有意に高まることを示唆している。

\section*{Discussion}
In this study, we examined the influence of volumetric motion, specifically expansion--contraction, on the perception of animacy in robots, using the physical robot MOFU. 
% In this study, we examined the influence of volumetric motion, specifically expansion-contraction, on the perception of animacy in robots, using the physical robot MOFU.
MOFU employs a Jitterbug mechanism, enabling whole-body expansion and contraction with a simple configuration, and its movements were used to evaluate animacy impressions. 
% MOFU employs a Jitterbug mechanism, which enables whole-body expansion and contraction with a simple configuration, and its movements were used to evaluate animacy impressions.
Here, we discuss the interpretation of the results, limitations of the study design, and directions for future research.
% Below, we discuss the interpretation of the results, limitations of the study design, and future research directions.

The results of this study indicate that both expansion--contraction (EC) and rotational motion (RM) in the Jitterbug structure significantly influenced animacy perception, while their interaction was not significant. 
% The results of this study showed that both expansion-contraction (EC) and rotational motion (RM) in the Jitterbug structure had significant main effects on animacy perception, whereas their interaction was not significant.
Post hoc comparisons showed no significant difference between the EC and RM+EC conditions, suggesting that adding RM does not provide additional benefit when EC is present. 
% Post hoc comparisons further revealed no significant difference between the EC and RM+EC conditions, indicating no additive benefit of adding RM when EC is present.
Although the RM+EC condition produced higher animacy ratings than RM alone, direct comparisons did not show that EC surpassed RM. 
% At the same time, the RM+EC condition exceeded RM, and we did not find evidence that EC exceeded RM in the direct comparison; therefore, we refrain from claiming dominance.
Overall, both movements enhanced animacy perception, with EC alone being sufficient to elicit a sense of animacy in this paradigm.
% In summary, both movements enhanced animacy perception, and EC appears sufficient to elicit animacy in our paradigm.

%The results of this study showed that both expansion-contraction (EC) and rotational movements (RM) in the Jitterbug structure had significant main effects on animacy perception, whereas their interaction was not significant. This indicates that EC is a particularly strong factor in eliciting animacy in robots. Post hoc comparisons further revealed no significant difference between the EC and RM+EC conditions, suggesting that adding RM does not increase animacy when EC is already present. In summary, although both movements enhanced animacy perception, EC emerged as the more critical determinant.

In the multi-robot conditions, NBM (2 robots) consistently received lower animacy ratings than the other conditions (RM, EC, RM+EC). 
% In the multi-robot conditions, NBM (2 robots) consistently received lower animacy ratings compared to other conditions (RM, EC, RM+EC).
This suggests that the ``type of movement,'' such as rotation or expansion--contraction, may have a stronger influence on perceived animacy than the number of robots. 
% This result suggests that the ``type of movement,'' such as rotation or expansion-contraction, may contribute more strongly to animacy perception than the number of robots. 
Although we hypothesized that presenting two robots simultaneously would increase animacy ratings compared with a single robot, this prediction was not supported, as no significant difference was observed between single- and dual-robot presentations. 
% Although we hypothesized that presenting two robots simultaneously would lead to higher animacy ratings compared with a single robot, this prediction was not supported, as the results showed no significant difference between the single- and dual-robot conditions.
Single-robot presentation may focus attention on ``individual movement,'' whereas multi-robot presentation could raise expectations for ``inter-robot relationships'' or ``interactions.'' 
% While single-robot presentation may direct attention toward ``individual movement,'' multi-robot presentation may raise expectations for ``inter-robot relationships'' or ``interactions.''
However, in this study, no elements suggesting coordination or competition between robots were designed, and the movement sequences were set independently.
% Nevertheless, in this study, no elements suggesting coordination or competition between robots were designed, and the movement sequences were set independently.

In addition, animacy ratings were higher when expansion--contraction was combined with locomotion than when locomotion occurred alone. 
% In addition, animacy ratings were higher when expansion-contraction was combined with locomotion than when locomotion occurred alone.
Parovel et al. \cite{Parovel2018-xh} reported that moving squares with area changes, such as caterpillar-like deformation, were perceived as more animate than squares undergoing simple linear motion. 
% Parovel et al. \cite{Parovel2018-xh} reported that moving squares accompanied by area changes, such as caterpillar-like deformation, were perceived as more animate compared with simple linear motion.
As MOFU also generates volumetric changes through expansion and contraction, these findings imply that such movements could be an effective way to add animacy to locomotion in real-world robotic settings.
% Since MOFU also produces volumetric changes through expansion-contraction, these findings suggest that such movements may serve as an effective means of adding animacy to locomotion in real-world robotic settings.

\subsection*{Limitations}
However, this study has certain limitations. 
First, it focused on the presence or absence of expansion--contraction but did not directly compare it with other movement types, such as articulated or axial motion. 
% First, this study focused on the presence or absence of expansion-contraction but did not include direct comparisons with other movement types (e.g., articulated motion or axial motion).
Therefore, the specific contribution of expansion--contraction to enhancing animacy relative to other movement forms remains unclear. 
% Therefore, it remains unclear to what extent expansion-contraction specifically enhances animacy compared with other forms of movement. 
Additionally, parameters such as the cycle and amplitude of expansion--contraction, as well as MOFU’s external appearance, were not systematically examined.
% Moreover, parameters such as the cycle and amplitude of expansion-contraction, as well as the external appearance of MOFU, were not examined.

Second, the multi-robot conditions only increased the number of robots presented, without introducing interactions between them. 
% Second, the multi-robot conditions simply increased the number of robots presented, without incorporating situations where the robots interacted with each other. 
Future research should explore coordinated or competitive behaviors, which may elicit higher levels of perceived animacy.
% Future work should investigate coordinated or competitive behaviors between robots, which may elicit higher levels of animacy.

Third, while combining locomotion with expansion--contraction was shown to enhance animacy, the effects of synchronization and motion parameters, such as speed, timing, and smoothness, were not analyzed. 
% Third, although combining locomotion with expansion-contraction was shown to be effective, the influence of synchronization, as well as motion parameters such as speed, timing, and smoothness, on animacy ratings was not analyzed.
Precise control of these parameters could help identify the underlying perceptual mechanisms.
% More precise control of these parameters may help identify the underlying perceptual factors.

Fourth, MOFU was covered with a fur-like exterior to promote a natural appearance and potential physical interaction; however, the specific impact of appearance on animacy perception was not assessed. 
% Fourth, MOFU was covered with a fur-like exterior to consider potential future physical interaction and a natural appearance, but the influence of appearance itself was not assessed.
Evaluations involving direct interaction with humans remain necessary.
% In addition, evaluations of animacy in contexts involving interaction with humans remain necessary.

\subsection*{Future directions}
The findings of this study suggest that fundamental volumetric motion, such as expansion--contraction, can independently evoke a sense of animacy, even without locomotion. 
% The findings of this study suggest that fundamental volumetric motion, such as expansion-contraction, can independently elicit animacy even in the absence of locomotion.
This indicates that expansion--contraction alone may serve as an effective design feature for robots in medical and welfare contexts, where space is limited and quiet operation is often essential.
% This indicates that expansion-contraction alone may serve as an effective design feature for robots used in medical and welfare contexts, where limited space and quiet operation are often required.

Furthermore, investigating expansion--contraction as a fundamental form of three-dimensional motion could pave the way for incorporating diverse volumetric changes observed in human gestures and animal behaviors into robotic designs. 
% Furthermore, exploring expansion-contraction as a basic principle of three-dimensional motion could open the way to applying diverse volumetric changes observed in human gestures and animal behaviors to robots.
This approach may ultimately provide guidelines for creating robots capable of more intuitive and expressive interactions.

\section*{Conclusion}
In this study, we investigated the effect of expansion--contraction movements (volumetric motion) on the perception of animacy in robots using a video-based survey with the physical robot MOFU (MOrphing Fluffy Unit). 
% In this study, we examined the effect of expansion-contraction movements (volumetric motion) on the perception of animacy in robots through a video-based survey using the physical robot MOFU (Morphing Fluffy Unit). 
MOFU features a Jitterbug structure that enables whole-body expansion and contraction driven by a single linear actuator and is also equipped with a differential two-wheel drive for locomotion.
% MOFU was designed with a Jitterbug structure that enables whole-body expansion and contraction driven by a single degree of freedom linear actuator, and it is also equipped with a differential two-wheel drive for locomotion.

The experimental results showed, first, that both expansion--contraction and rotational motion significantly increased animacy ratings compared with a baseline without body motion, with no significant difference between them in the direct comparison. 
% The experimental results showed, first, that both expansion-contraction and rotational motion significantly increased animacy ratings relative to a baseline without body motion, and that there was no significant difference between them in the direct comparison.
Second, presenting two robots did not significantly increase animacy relative to presenting a single robot under the expansion--contraction condition, suggesting that simply increasing the number of robots does not necessarily enhance animacy. 
% Second, presenting two robots did not significantly increase animacy relative to presenting a single robot under the expansion-contraction condition, suggesting that simply increasing the number of robots may not necessarily enhance animacy.
Conditions without movement, including those lacking rotational and volumetric motion, were consistently rated lower, highlighting the importance of the ``type of movement'' in animacy perception. 
% However, conditions without movement (i.e., without rotational and volumetric motion) were consistently rated lower, confirming the importance of the ``type of movement'' in animacy perception.
Third, combining expansion--contraction with locomotion significantly increased animacy ratings compared with locomotion alone. 
% Third, combining expansion-contraction with locomotion significantly increased animacy ratings compared with locomotion alone.
All experiments employed between-participants comparisons, meaning the findings reflect consistent tendencies across independent participant groups rather than direct within-subject contrasts.
% It should be noted that all experiments in this study were conducted with between-participants comparisons, and thus the findings reflect consistent tendencies observed across independent participant groups rather than direct within-subject contrasts.

These findings indicate that relatively simple volumetric changes, such as expansion and contraction, can effectively enhance the perception of animacy without the need for complex appearances or mechanisms.

\section*{Acknowledgments}
This work was carried out as part of a joint research project, with the support and collaboration of Sony Corporation.

\nolinenumbers

\bibliography{reference}

\end{document}